%% file: main.tex
\documentclass[11pt,a4paper]{article}

\usepackage[margin=23mm,top=25mm,bottom=25mm,headheight=15pt]{geometry}
\usepackage{fontspec}
\usepackage{amsmath,amssymb,amsthm,mathtools}
\usepackage{tikz}
\usetikzlibrary{arrows.meta,positioning,calc}
\usepackage{forest}
\useforestlibrary{edges}
\usepackage{bm}
\usepackage{booktabs,array,calc,multirow,longtable}
\usepackage{graphicx}
\usepackage{enumitem}
\usepackage{microtype}
\usepackage[numbers,sort&compress]{natbib}
\usepackage{xcolor}
\definecolor{linkblue}{HTML}{175A88}
\usepackage{xurl}
\usepackage[colorlinks=true,linkcolor=linkblue,citecolor=linkblue,urlcolor=linkblue,
  pdftitle={ Trajectory Density Ratios under Weak Support: A Survey across the Sciences },
  pdfsubject={Weak support in trajectory density-ratio estimation: levels, retreats, and cross-domain applications}]{hyperref}
\usepackage{bookmark}
\usepackage{fancyhdr}

\fancypagestyle{plain}{%
  \fancyhf{}%
  \fancyfoot[C]{\thepage}%
}

\theoremstyle{plain}
\newtheorem{observation}{Observation}
\newtheorem{proposition}{Proposition}
\theoremstyle{remark}

\let\surveyEqref\eqref
\input{math_commands.tex}

\let\eqref\surveyEqref

\newcommand{\Prob}{\mathbb{P}}
\newcommand{\dd}{\,\mathrm{d}}
\newcommand{\esssup}{\operatorname{ess\,sup}}
\newcommand{\TV}{\mathrm{TV}}

\renewcommand{\arraystretch}{1.12}
\setlist{itemsep=0.2em,topsep=0.3em,leftmargin=2em}

\title{\bfseries Trajectory Density Ratios under Weak Support:\\ A Survey across the Sciences}
\author{Wei Chen, Lihao Zhuang, Yubing Wu, and Delu Zeng\\[0.4em]
\normalsize South China University of Technology, Guangzhou, China}
\date{}

\begin{document}
\maketitle
\thispagestyle{plain}

\begin{abstract}
\noindent Many problems across the sciences compare two stochastic processes through the sequences they generate.
Reinforcement learning asks what a new policy would return from the trajectories of an old one; molecular simulation asks how much more a ligand binds from samples of the unbound state; climate attribution asks how a heat wave shifted under warming; and language-model post-training asks what a deployed agent will do from logged interactions.
In every case, one quantity would answer all these questions at once: the ratio of the two probability laws over their generated sequences, the trajectory density ratio.
This ratio exists only under strong conditions.
When one process reaches states the other cannot, the ratio does not exist; when their overlap is thin, its variance grows without control with the sequence length; and when they concentrate on different lower-dimensional sets, the ratio is undefined altogether.
Weak support of this kind is the norm for sequence data, not a sampling accident, and its price is paid in identifiability rather than in noise alone.
This survey organizes the field around a three-level diagnosis of weak support and a map of the nine retreats that methods take when the full ratio is out of reach, from direct estimation and reweighting through staged decomposition, stationarization, and partial-identification bounds to smoothing, transport couplings, operator spectra, and direct estimation of a single functional.
Each retreat keeps part of what the ratio offers and pays for the rest, and each application domain has standardized the retreat its data force.
The survey closes with open problems: testable diagnostics at every level, partial identification at the trajectory level, a rate theory of time-amplified error, and benchmarks with ground truth.

\end{abstract}

\clearpage
\tableofcontents
\clearpage

\input{sections/survey_intro}
\input{sections/survey_formalization}
\input{sections/survey_r1}
\input{sections/survey_r2}
\input{sections/survey_r3}
\input{sections/survey_r4}
\input{sections/survey_r56}
\input{sections/survey_r7}
\input{sections/survey_r8}
\input{sections/survey_domains}
\input{sections/survey_open}

\clearpage
\begingroup
\small
\bibliographystyle{unsrtnat}
\bibliography{references}
\endgroup

\clearpage
\appendix
\input{sections/survey_appendix}
\input{sections/survey_retreatmap}

\end{document}

%% file: math_commands.tex
\usepackage{amsmath,amsfonts,bm}

\def\eqref#1{equation~\ref{#1}}

\def\1{\bm{1}}

\def\rmA{{\mathbf{A}}}

\def\rmX{{\mathbf{X}}}

\def\rmZ{{\mathbf{Z}}}

\def\va{{\bm{a}}}

\def\vh{{\bm{h}}}

\def\vs{{\bm{s}}}

\def\vx{{\bm{x}}}

\DeclareMathAlphabet{\mathsfit}{\encodingdefault}{\sfdefault}{m}{sl}
\SetMathAlphabet{\mathsfit}{bold}{\encodingdefault}{\sfdefault}{bx}{n}

\def\gA{{\mathcal{A}}}

\def\gF{{\mathcal{F}}}

\def\gX{{\mathcal{X}}}

\newcommand{\E}{\mathbb{E}}



%% file: sections/survey_intro.tex
\section{Introduction}\label{sec:intro}

Many scientific questions take the same form: we hold trajectories generated by one process, but the
question of interest concerns the trajectory distribution of another process. Off-policy evaluation
in reinforcement learning asks what return a target policy would achieve while the logs come from a
behavior policy \citep{liu2018horizon,nachum2019dualdice,thomas2015high}. Computational chemistry
compares free energies at two temperatures or two ligand parameters while samples come from
molecular dynamics or umbrella sampling \citep{kirkwood1935coupled,kumar1992weighted,bennett1976}.
Single-cell biology reconstructs differentiation from destructive sampling \citep{trapnell2014dynamics,saelens2019comparison}.
Climate attribution estimates the probability ratio of the same extreme event under factual and
counterfactual forcing \citep{stott2004human,otto2017attribution}. Ecology compares locations that
animals use with locations that are available \citep{johnson1980comparison,lele2006weighted}, and
the post-training of large language models reuses responses logged under earlier policies to
evaluate and improve a new one \citep{ouyang2022training,rafailov2023dpo,yao2024taubench}. The
objects, data, and vocabularies differ. The mathematical structure does not. A data process $Q$
supplies the samples, a target process $P$ defines the question, and the bridge is the
Radon--Nikodym derivative $\dd P/\dd Q$ between trajectory measures: the trajectory density ratio, or its
logarithm, the trajectory log-ratio.

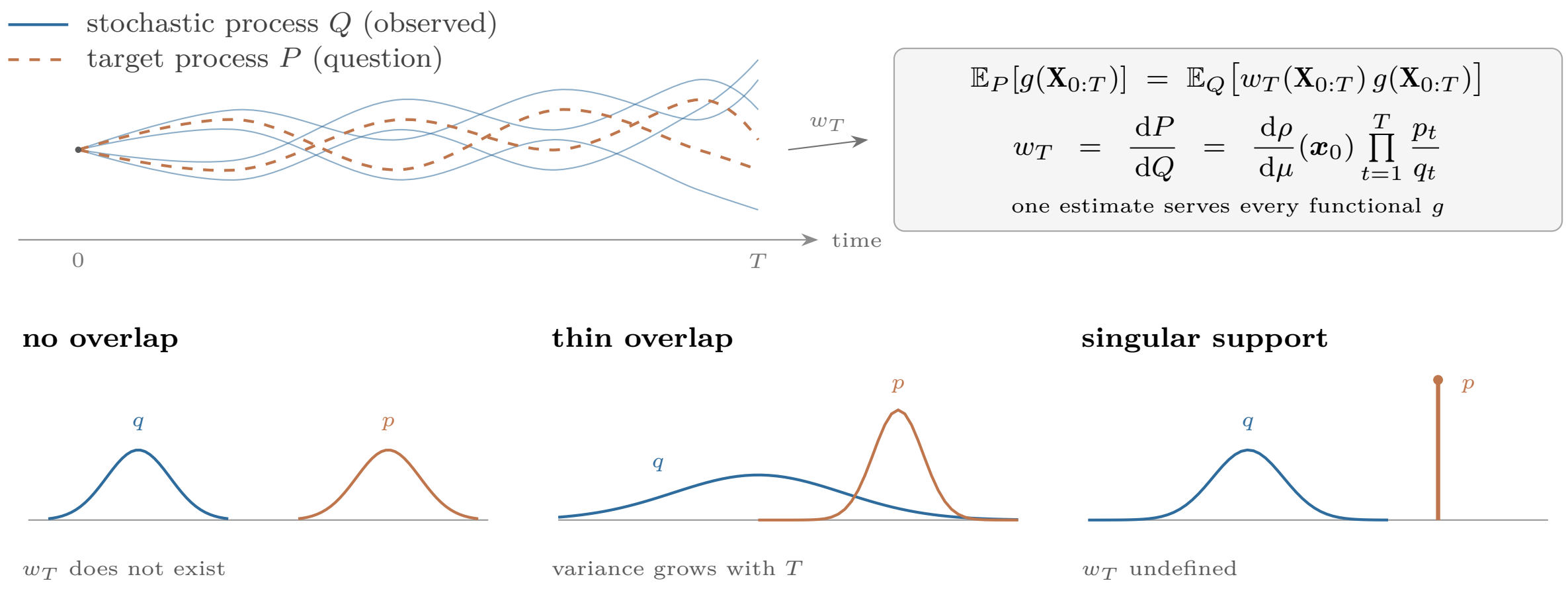
\begin{figure}[ht]
\centering
\input{figures/fig_problem}
\caption{The trajectory-density-ratio problem. Top: we observe trajectories of a stochastic
process $Q$ but need expectations under a target process $P$; the ratio $w_T=\dd P/\dd Q$, a
product of one-step ratios, converts one into the other. Bottom: weak support breaks this bridge
in three ways---no overlap leaves $w_T$ undefined, thin overlap makes its variance grow with the
horizon, and singular support leaves no density.}
\label{fig:problem}
\end{figure}

Trajectory density ratios differ from their independent and identically distributed counterparts in
two structural ways. First, they factorize along time. For a Markov process the trajectory ratio is the
product of an initial ratio and the one-step transition-density ratios, so the log-ratio is a sum
of local increments (Section~\ref{sec:formalization}). Second, support conditions multiply along
time. If at a single time step the target process enters, with positive probability, a region the
data process cannot reach, the ratio of the whole trajectory ceases to exist: a local support gap
escalates, under the product structure, to singularity of the trajectory measures. The same product
structure amplifies thinness of overlap: the second moment of the ratio grows exponentially with
the horizon whenever one-step ratios have non-degenerate tails. Weak support is therefore the
operative regime of process data, not a finite-sample nuisance that more data can remove.
Figure~\ref{fig:problem} draws the object and the three ways it fails.

\paragraph{Core claim.}
The claim that organizes this survey is best stated directly. \textbf{The trajectory ratio $w_T=\dd P/\dd Q$
is the unique estimand that a single estimate reuses across every trajectory functional, and the
literature on trajectory density ratios is a map of \emph{retreats} from this object: each method
family computes, instead of $w_T$, a surrogate estimand adopted because weak support makes $w_T$
nonexistent, uncontrollable, or singular.} The families differ not in
ambition but in what their surrogate keeps and what it pays. Truncation keeps the object
and pays in tail bias. Stationarization keeps long-run functionals and gives up the trajectory.
Partial identification keeps validity and gives up point estimates. Couplings and operator spectra
give up ratio values altogether. The application domains differ in which retreat their data force
them to standardize. Weak support is the pressure. The map of retreats (Figure~\ref{fig:retreats},
Table~\ref{tab:retreats}) is the organization, and everything else in this survey is a branch of it.

Two research traditions have each solved half of the problem. Statistics and machine learning
developed density-ratio estimation as a learning problem: convex and least-squares variational
objectives \citep{sugiyama2007direct,kanamori2009ulsif,nguyen2010convex,sugiyama2012bregman},
kernel two-sample machinery \citep{gretton2012mmd,muandet2017kernel}, convergence rates for
regularized regression \citep{caponnetto2007optimal,tsybakov2009introduction}, and a unifying link
between classification losses and ratio estimation \citep{menon2016,gutmann2012}. Off-policy
evaluation refined this into a family of estimators organized around stationary distribution
corrections \citep{liu2018horizon,nachum2019dualdice,zhang2020gendice,yang2020off}, together with a
theory of what remains identifiable when coverage fails \citep{wang2021statistical,foster2022offline,jin2021pessimism,khan2023off}.
Computational chemistry, dynamical systems, and climate science have lived with rare trajectories
for far longer and accumulated a different toolbox: phase-space overlap measures
\citep{wu2005phase,wu2005relative}, large-deviation theory \citep{touchette2009large,dembo1998large},
transfer operators \citep{lasota1994chaos,baladi2000positive}, and rare-event algorithms
\citep{ragone2017computation,bouchet2019rare}. Neither tradition states the connection the other
way around: the OPE estimators are retreats to marginals and functionals, and the physics toolbox
is a catalog of the same retreats under different names. Making the retreat explicit is what lets
the two traditions be compared at all.

The survey carries this claim through in four steps. First, it formalizes the object and the
pressure: the product structure of $w_T$, the three levels of weak support, the two escalation
mechanisms (Observations~\ref{obs:escalation} and~\ref{obs:amplification}), and the
identifiability--rate consequences of each level. It then defines a retreat precisely and draws the
map (Section~\ref{sec:formalization}). Second, the body of the survey is organized \emph{by the
retreat itself}, one subsection per row of the map (Section~\ref{sec:tour}): direct estimation and
reweighting stay on the object (\S\ref{sec:reweighting}), staged and sequential decomposition
splits the product (\S\ref{sec:sequence}), stationarization trades the trajectory for the horizon
(\S\ref{sec:dice}), partial identification bounds what coverage admits
(\S\ref{sec:partial}), bridges and transport change the estimand (\S\ref{sec:bridges}),
operator spectra replace densities (\S\ref{sec:operators}), and direct functionals and model
substitution bypass the ratio altogether (\S\ref{sec:bypass}). Each subsection closes with the
domains that standardized the retreat, so the application literature is read through the map
rather than listed beside it. Third, Section~\ref{sec:domains} juxtaposes the estimator families, states the
cross-domain comparison keyed by retreat, and reads the newest domain, language-model
post-training with multi-turn agents, as a composite in which nearly every retreat operates at
once. Fourth, it recasts the open problems as prescriptions for retreating later and returning
sooner (Section~\ref{sec:open}).

The comparison across domains rests on the retreat grid rather than on shared theorems. What one
field calls phase-space overlap another calls the effective sample size, a third calls positivity,
and a fourth calls concentrability: all of them measure how close the problem sits to the level
that forces a retreat. The value of the grid is practical. A domain that has standardized a
diagnostic can adopt the matching estimator family without re-deriving its failure modes, and a
method's known pathologies follow from its retreat rather than from implementation quality.

\paragraph{Reader's guide.}
Four artifacts support navigation. Figure~\ref{fig:retreats} (Section~\ref{sec:retreats}) draws the
map of retreats that carries the core claim, and Table~\ref{tab:retreats}
(Appendix~\ref{app:retreats}) states it on one page, with every retreat's level, what its surrogate keeps and what it pays, and
its standardizing domains. Figure~\ref{fig:timeline} (Section~\ref{sec:sequence}) places the two
research lineages on a common time axis. Table~\ref{tab:methods} and Table~\ref{tab:domains}
(Section~\ref{sec:domains}) contrast the estimator families and the retreat-keyed cross-domain
comparison, and Table~\ref{tab:tensions} (Section~\ref{sec:seq-inference}) distills the three
tensions of the sequential literature. A reader looking for where a specific tool sits should
start from the map and follow the section pointers back into the text.

This is a thematic review, not a systematic one. The corpus consists of two curated
bibliographies, namely a sequence-level density-ratio corpus maintained alongside this survey and a
weak-support corpus assembled for Section~\ref{sec:formalization}, supplemented by the canonical
literature of each application domain. Selection follows three threads: density ratio, weak
overlap, and process trajectories. The scope limits completeness within domains but makes the
concepts of the different traditions alignable entry by entry.

%% file: figures/fig_problem.tex
\definecolor{probQ}{HTML}{2F6C9E}
\definecolor{probP}{HTML}{C0764C}
\begin{tikzpicture}[font=\scriptsize, x=1mm, y=1mm, >={Stealth}]

\draw[probQ, thick] (3,58.5) -- (9,58.5);
\node[anchor=west, black!75] at (9.5,58.5) {stochastic process $Q$ (observed)};
\draw[probP, thick, dashed] (3,55) -- (9,55);
\node[anchor=west, black!75] at (9.5,55) {target process $P$ (question)};

\draw[black!45, -{Stealth}] (4,37) -- (84,37) node[right, font=\tiny, black!60] {time};
\node[font=\tiny, black!55, below] at (10,37) {$0$};
\node[font=\tiny, black!55, below] at (78,37) {$T$};

\draw[probQ, opacity=0.55, smooth, tension=0.6] plot coordinates
  {(10,46) (26,50) (42,47) (58,52) (72,49) (78,53)};
\draw[probQ, opacity=0.55, smooth, tension=0.6] plot coordinates
  {(10,46) (26,43) (42,48) (58,44) (72,50) (78,55)};
\draw[probQ, opacity=0.55, smooth, tension=0.6] plot coordinates
  {(10,46) (26,48) (42,43) (58,47) (72,42) (78,40)};
\draw[probQ, opacity=0.55, smooth, tension=0.6] plot coordinates
  {(10,46) (26,45) (42,51) (58,48) (72,53) (78,50)};
\draw[probP, thick, dashed, smooth, tension=0.6] plot coordinates
  {(10,46) (26,44) (42,49) (58,46) (72,51) (78,47)};
\draw[probP, thick, dashed, smooth, tension=0.6] plot coordinates
  {(10,46) (26,49) (42,44) (58,50) (72,46) (78,44)};
\fill[black!65] (10,46) circle (0.9pt);

\draw[-{Stealth}, black!55] (81,46) -- (89,47)
  node[midway, above, font=\tiny] {$w_T$};

\node[draw=black!35, rounded corners=3pt, fill=black!4, align=center,
      text width=64mm, inner sep=4pt] at (125,47) {%
  $\E_P[g(\rmX_{0:T})]=\E_Q\bigl[w_T(\rmX_{0:T})\,g(\rmX_{0:T})\bigr]$\\[3pt]
  $w_T=\dfrac{\dd P}{\dd Q}=\dfrac{\dd\rho}{\dd\mu}(\vx_0)\prod\limits_{t=1}^{T}\dfrac{p_t}{q_t}$\\[3pt]
  {\tiny one estimate serves every functional $g$}};

\draw[black!40] (5,9) -- (51,9);
\draw[probQ, thick] plot[domain=7:25, samples=40]
  (\x, {9+7*exp(-((\x-16)/4.5)*((\x-16)/4.5))});
\draw[probP, thick] plot[domain=32:50, samples=40]
  (\x, {9+7*exp(-((\x-41)/4.5)*((\x-41)/4.5))});
\node[probQ, font=\tiny] at (16,18.6) {$q$};
\node[probP, font=\tiny] at (41,18.6) {$p$};
\node[anchor=west, font=\scriptsize\bfseries] at (3,27) {no overlap};
\node[anchor=west, font=\tiny, black!65] at (3,4) {$w_T$ does not exist};

\draw[black!40] (58,9) -- (104,9);
\draw[probQ, thick] plot[domain=58:104, samples=60]
  (\x, {9+4.5*exp(-((\x-78)/12)*((\x-78)/12))});
\draw[probP, thick] plot[domain=78:104, samples=40]
  (\x, {9+11*exp(-((\x-92)/3.5)*((\x-92)/3.5))});
\node[probQ, font=\tiny] at (68,14.5) {$q$};
\node[probP, font=\tiny] at (92,22.4) {$p$};
\node[anchor=west, font=\scriptsize\bfseries] at (56,27) {thin overlap};
\node[anchor=west, font=\tiny, black!65] at (56,4) {variance grows with $T$};

\draw[black!40] (111,9) -- (157,9);
\draw[probQ, thick] plot[domain=111:141, samples=50]
  (\x, {9+7*exp(-((\x-127)/5)*((\x-127)/5))});
\draw[probP, very thick] (146,9) -- (146,23);
\fill[probP] (146,23) circle (1.4pt);
\node[probQ, font=\tiny] at (127,18.6) {$q$};
\node[probP, font=\tiny] at (149,22.4) {$p$};
\node[anchor=west, font=\scriptsize\bfseries] at (109,27) {singular support};
\node[anchor=west, font=\tiny, black!65] at (109,4) {$w_T$ undefined};

\end{tikzpicture}

%% file: sections/survey_formalization.tex
\section{Trajectory density ratios and weak support}\label{sec:formalization}

\subsection{Processes, trajectories, and target functionals}\label{sec:processes}

Let $\gX$ be the state space and $\rmX_{0:T}=(\rmX_0,\dots,\rmX_T)$ a length-$T$ sequence, that is, a trajectory (an observed time series when $t$ indexes time), taking values in the trajectory space $\Omega_T=\gX^{T+1}$, the space of all length-$T$
sequences, equipped with the natural filtration
$\gF_T=\sigma(\rmX_0,\dots,\rmX_T)$. The data process is generated by an initial distribution $\mu$ and
transition kernels $q_t(\cdot\mid \vx_{t-1})$. Its trajectory measure is $Q$. The target process is
generated by $\rho$ and kernels $p_t(\cdot\mid \vx_{t-1})$. Its trajectory measure is $P$. In off-policy
evaluation the two differ only in the policy \citep{puterman1994markov,sutton2018reinforcement}.
In free-energy calculation they differ in the Hamiltonian or the biasing potential
\citep{kirkwood1935coupled,torrie1977nonphysical}. In climate attribution they correspond to
factual and counterfactual forcing \citep{stott2004human,otto2017attribution}. In ecology they
correspond to the availability and use sampling mechanisms \citep{johnson1980comparison,aarts2013quantifying}.
In language-model post-training they correspond to the reference and the trained policy, or to two
successive versions of it \citep{schulman2017proximal,rafailov2023dpo}. The common structure is:
we observe samples of $Q$ and must answer questions about $P$.

The target quantity is usually an expectation of a trajectory functional, such as a discounted return
$\E_P[\sum_{t=0}^{T}\gamma^t r(\rmX_t,\rmA_t)]$, a finite-horizon hitting probability, or the free-energy
difference between the two processes. We restrict attention to quantities writable as
$\E_P[g(\rmX_{0:T})]$. More complex functionals (quantiles, conditional value-at-risk, causal
counterfactual effects beyond the change-of-measure form) can be handled pointwise in the same
framework but require additional continuity arguments \citep{vandervaart1998}.

\subsection{The trajectory ratio and its product structure}\label{sec:product}

If $P$ is absolutely continuous with respect to $Q$ on $\gF_T$, the trajectory density ratio
$w_T=\dd P/\dd Q$ exists and, for every $P$-integrable $g$,
\begin{equation}\label{eq:change-of-measure}
  \E_P\bigl[g(\rmX_{0:T})\bigr] \;=\; \E_Q\bigl[w_T(\rmX_{0:T})\, g(\rmX_{0:T})\bigr].
\end{equation}
For Markov processes the Radon--Nikodym derivative factorizes,
\begin{equation}\label{eq:product}
  w_T(\vx_{0:T}) \;=\; \frac{\dd\rho}{\dd\mu}(\vx_0)\,\prod_{t=1}^{T}\frac{p_t(\vx_t\mid \vx_{t-1})}{q_t(\vx_t\mid \vx_{t-1})},
\end{equation}
and in logarithmic form the trajectory log-ratio $L_T=\log w_T$ is a sum of one-step log-ratios
\citep{girsanov1960measures,karatzas1991brownian,revuz1999continuous}. Equation~\eqref{eq:product}
is the starting point reused throughout this survey. It says that the trajectory ratio is not a new
estimation target but a product of one-step ratios. It also says that any defect of a one-step ratio
is compounded multiplicatively. In particular, if a single transition has zero density under the
data process and positive density under the target process, then equation~\eqref{eq:product} is
undefined on that trajectory and equation~\eqref{eq:change-of-measure} fails with it.

When the process is stationary and only long-run averages matter, the product in
equation~\eqref{eq:product} can be replaced by a stationary marginal ratio. Let $\dd\pi$ and
$\dd D$ be the stationary state--action distributions of the target and data processes. The
stationary ratio $\zeta=\dd\pi/\dd D$ satisfies $\E_{\pi}[h]=\E_D[\zeta h]$ independently of the
horizon. The DICE family is built on this substitution \citep{liu2018horizon,nachum2019dualdice,yang2020off}.
Section~\ref{sec:dice} discusses how it trades the product-level difficulty for a coverage and
mixing problem at the stationary level \citep{meyn2009markov,norris1997markov}.

Two related but distinct objects should not be conflated with $w_T$. Fixed-window ratios compare
finite segments and retain only short-range order \citep{kawahara2012changepoint,liu2013rulsif}.
Relative and mixture-denominator ratios trade unboundedness for stability and change the estimand
\citep{yamada2013relative,sugiyama2012bregman}. Conditional prefix ratios
$L_t=\log\frac{p(\vx_t|\vx_{<t})}{q(\vx_t|\vx_{<t})}$ preserve trajectory information incrementally but accumulate
estimation error additively in the log scale
(Section~\ref{sec:sequence}), and trajectory-level information functionals such as the relative entropy
rate target divergences rather than pointwise ratios
\citep{pantazis2013rer,dupuis2016pathbounds,cover2006}.

\subsection{Three levels of weak support}\label{sec:levels}

We read weak support as a failure of the support \emph{relation} between $P$ and $Q$ rather than as
a topological property of either distribution alone. Strictly, ``weak support set'' describes a
ternary relation: whether the support is contained, whether the overlap is thick enough, and
whether the trajectory measures are absolutely continuous. We distinguish three diagnostic levels.

\paragraph{$L_1$: support non-coverage.} There exists an $\gF_T$-measurable set $\gA$ with
$\Prob_P(\gA)>0$ and $\Prob_Q(\gA)=0$. Equivalently, $P$ is not absolutely continuous with respect to
$Q$, and the trajectory density ratio does not exist. At the one-step level this appears when a
target-reachable transition has zero probability under the data process:
$p_t(\gA\mid \vx)>0$ while $q_t(\gA\mid \vx)=0$. In off-policy evaluation this occurs when the behavior
policy assigns zero probability to certain actions, a condition known as deficient support
\citep{sachdeva2020off,thomas2015high}. In causal inference it is a positivity violation
\citep{westreich2010invited,petersen2010diagnosing,cole2008constructing}. In ecology it occurs when
the availability definition does not cover the habitats actually used
\citep{aarts2013quantifying,northrup2013practical}. In free-energy calculation it is the
non-overlap of adjacent umbrella windows \citep{wu2005phase,torrie1977nonphysical}. Truncated and
censored data models study precisely this regime from the estimation side
\citep{daskalakis2018efficient}.

\paragraph{$L_2$: thin overlap.} $P$ is absolutely continuous with respect to $Q$ but the ratio is
unbounded: $\esssup w_T=\infty$. If moreover $\E_Q[w_T^2]<\infty$, we call the overlap
\emph{square-integrably weak}. The ratio exists with finite second moment, but the trade-off between
truncation and variance becomes the central object \citep{mehrabi2024off}. If
$\E_Q[w_T^2]=\infty$, the ratio remains integrable but the variance of any estimator escapes
uniform control. Strong overlap, $\esssup w_T\le M$, is one endpoint of this spectrum and
corresponds to the concentrability conditions of batch reinforcement learning
\citep{chen2019information,munos2016safe}. The thickness of the overlap can be measured by the
overlap coefficient
\begin{equation}\label{eq:overlap-coefficient}
  \varrho(P,Q) \;=\; \int \min(w_T,1)\,\dd Q \;=\; 1-\TV(P,Q),
\end{equation}
or by the R\'enyi-2 divergence
\begin{equation}\label{eq:renyi2}
  R_2(P\Vert Q) \;=\; \log \E_Q\bigl[w_T^2\bigr],
\end{equation}
which turns square-integrability into an estimable quantity and links ``is the overlap thick
enough'' directly to the second moment of the estimator \citep{cover2006,gretton2012mmd}. The
density-chasm regime, in which the endpoints are so far apart that the informative region carries
exponentially small mass under both laws, is the $L_2$ condition observed by classification-based estimators
\citep{rhodes2020tre,srivastava2023,albert1984separation}.

\paragraph{$L_3$: trajectory singularity.} $P$ and $Q$ are mutually singular on the trajectory space, or supported
on low-dimensional sets, so that no Radon--Nikodym derivative exists. This can occur while the
marginals overlap completely: two diffusions with different diffusion coefficients have mutually
singular trajectory measures under non-degeneracy \citep{girsanov1960measures,karatzas1991brownian,pavliotis2014stochastic};
orbits of deterministic chaotic systems are supported on attractors singular with respect to
Lebesgue measure \citep{lasota1994chaos,baladi2000positive}, and destructively sampled single-cell
trajectories do not correspond to the time evolution of a single cell
\citep{trapnell2014dynamics,saelens2019comparison}. $L_3$ can also arise as an upgrade of $L_1$.
This is the content of the next observation.

\begin{table}[t]
\centering\footnotesize
\caption{The three levels of weak support.}\label{tab:levels}
\setlength{\tabcolsep}{2.5pt}
\begin{tabular}{@{}>{\raggedright\arraybackslash}p{23mm}>{\raggedright\arraybackslash}p{20mm} *{6}{c} >{\raggedright\arraybackslash}p{24mm}@{}}
\toprule
Level & Condition &
\rotatebox{50}{\scriptsize supports overlap} &
\rotatebox{50}{\scriptsize low-dim.\ support} &
\rotatebox{50}{\scriptsize ratio exists} &
\rotatebox{50}{\scriptsize identifiable} &
\rotatebox{50}{\scriptsize consistent} &
\rotatebox{50}{\scriptsize worsens with horizon} &
Standard remedies \\
\midrule
$L_1$ support non-coverage &
$P\not\ll Q$ &
$\times$ & $\times$ & $\times$ & $\times$ & $\times$ & \checkmark &
restrict the space, report bounds \\
\addlinespace[8pt]
$L_2$ thin overlap &
$P\ll Q$, $\esssup w_T=\infty$ &
\checkmark & $\times$ & \checkmark & \checkmark & \checkmark & \checkmark &
truncation, stationarization \\
\addlinespace[8pt]
$L_3$ trajectory singularity &
$P\perp Q$ or low-dimensional support &
$\times$ & \checkmark & $\times$ & $\times$ & $\times$ & $\times$ &
smoothing, bridges, spectra \\
\bottomrule
\end{tabular}
\end{table}

\begin{observation}[Local gaps escalate to trajectory singularity]\label{obs:escalation}
Suppose the initial distributions are absolutely continuous and there exist a time $t$ and a set $\gA$
such that $p_t(\gA\mid \vx)>0$ while $q_t(\gA\mid \vx)=0$ for $\vx$ in a set of positive $\Prob_{t-1}$-measure
under $P$. Then $P$ and $Q$ are mutually singular on $\gF_T$.
\end{observation}

The proof is immediate from the cylinder-set representation: the probability of a cylinder set is a
product of transition-kernel probabilities, so a cylinder containing the forbidden transition has
positive $P$-mass and zero $Q$-mass. Along equation~\eqref{eq:product} the $Q$-density of that
cylinder contains a zero factor while the $P$-density is positive. Observation~\ref{obs:escalation}
explains why support problems in off-policy evaluation, free-energy calculation, and rare-event
simulation worsen with the time scale: not because errors accumulate, but because identifiability
itself disappears at the trajectory level. It also shows that $L_1$ and $L_3$ are not exclusive classes
but the same failure observed at different scales.

\begin{observation}[Horizon amplification of variance]\label{obs:amplification}
If the one-step increment ratios $r_t=p_t/q_t$ are independent and identically distributed under
$Q$, then
\begin{equation}\label{eq:amplification}
  1+\chi^2(P_T\Vert Q_T) \;=\; \E_Q\bigl[w_T^2\bigr] \;=\; \E_Q\bigl[r_1^2\bigr]^{T} \;=\;
  \bigl(1+\chi^2(P_1\Vert Q_1)\bigr)^{T}.
\end{equation}
\end{observation}

Even when single-step overlap is only mildly thin, the second moment of the trajectory ratio grows
exponentially with the horizon. For mixing Markov chains the exponential rate is governed by the
spectral radius of the associated transfer operator \citep{lasota1994chaos,baladi2000positive,lezaud1998chernoff};
for diffusions the variance of the Girsanov density grows with
$\int_0^T\|\Delta b\|^2\dd t$ \citep{girsanov1960measures,karatzas1991brownian}. This is the
measure-theoretic statement of the horizon curse and the motivation for stationarization
(Section~\ref{sec:dice}). The corresponding sample-size statement for importance sampling is
exponential in the Kullback--Leibler discrepancy \citep{chatterjee2018is,kahn1953methods,glynn1989importance,agapiou2017importance}.
The product form of equation~\eqref{eq:amplification} also implies that controlling the
\emph{tail} of the one-step ratio matters more than controlling its mean.

The three levels form a diagnostic progression rather than an exclusive classification: $L_1$ asks
whether the support is contained, $L_2$ whether the overlap is thick enough, $L_3$ whether the
trajectory measures are absolutely continuous. A concrete problem can exhibit several levels at once, and the
choice of method depends on which level dominates the current error. Table~\ref{tab:levels}
summarizes the correspondence. The two mechanisms behind the table are the escalation of a local
gap to trajectory singularity and the exponential amplification of thin overlap.

\subsection{Identifiability and rates}\label{sec:rates}

The three levels have distinct statistical consequences. Under $L_1$ the density ratio is not
identifiable and the target functional is only partially identified: what is identifiable is a pair
of bounds whose width is determined by structural assumptions. \citet{khan2023off} characterize,
under Lipschitz continuity of conditional means, the contribution of the non-overlap region to
off-policy values by a pair of linear programs, obtain sharp bounds, and prove that the convergence
rate is minimax optimal up to logarithmic factors. Under $L_2$ the ratio is identifiable but rates
degrade: \citet{mehrabi2024off} prove that when the stationary ratio is square-integrable, a
truncated doubly robust estimator recovers the $\tfrac1T$ rate, that estimation remains consistent
without square-integrability at a slower rate, and that the slower rate is minimax on MDP classes
defined only by mixing conditions. Under $L_3$ no consistent estimator of the density ratio exists.
Every estimator must either introduce structural assumptions or change the estimand by smoothing.
Minimax lower bounds for off-policy evaluation formalize this trichotomy from the sampling side
\citep{wang2021statistical,foster2022offline,chen2019information,duan2020minimax,uehara2021finite}.

Together these results show that the price of weak support is not a single variance constant but a
joint change of identifiability, rates, and estimation object. A common misdiagnosis treats $L_1$
as an $L_2$ problem with large variance: truncation can stabilize the numerics but cannot make an
unidentified point estimate identifiable. Conversely, treating $L_2$ as $L_1$ discards
identifiable information and produces bounds wider than necessary. The task of diagnosis is to
determine, before estimation, at which level the problem sits.

\subsection{One object, many estimands}\label{sec:retreats}

What makes $w_T$ worth estimating is reusability: a single estimate of the change-of-measure
device~\eqref{eq:change-of-measure} answers every question of the form $\E_P[g(\rmX_{0:T})]$,
including returns, hitting probabilities, free-energy differences, and likelihood ratios between mechanisms.
and composes across horizons through the product structure~\eqref{eq:product}. This reusability is
also what makes its failure expensive. We call a \emph{retreat} any surrogate estimand that a
method computes instead of $w_T$. A retreat is characterized by three choices: the
\emph{trigger}, namely the level at which $w_T$ becomes nonexistent, uncontrollable, or singular;
what it \emph{keeps}, namely which questions the surrogate still answers; and what it \emph{pays},
namely which questions it gives up.

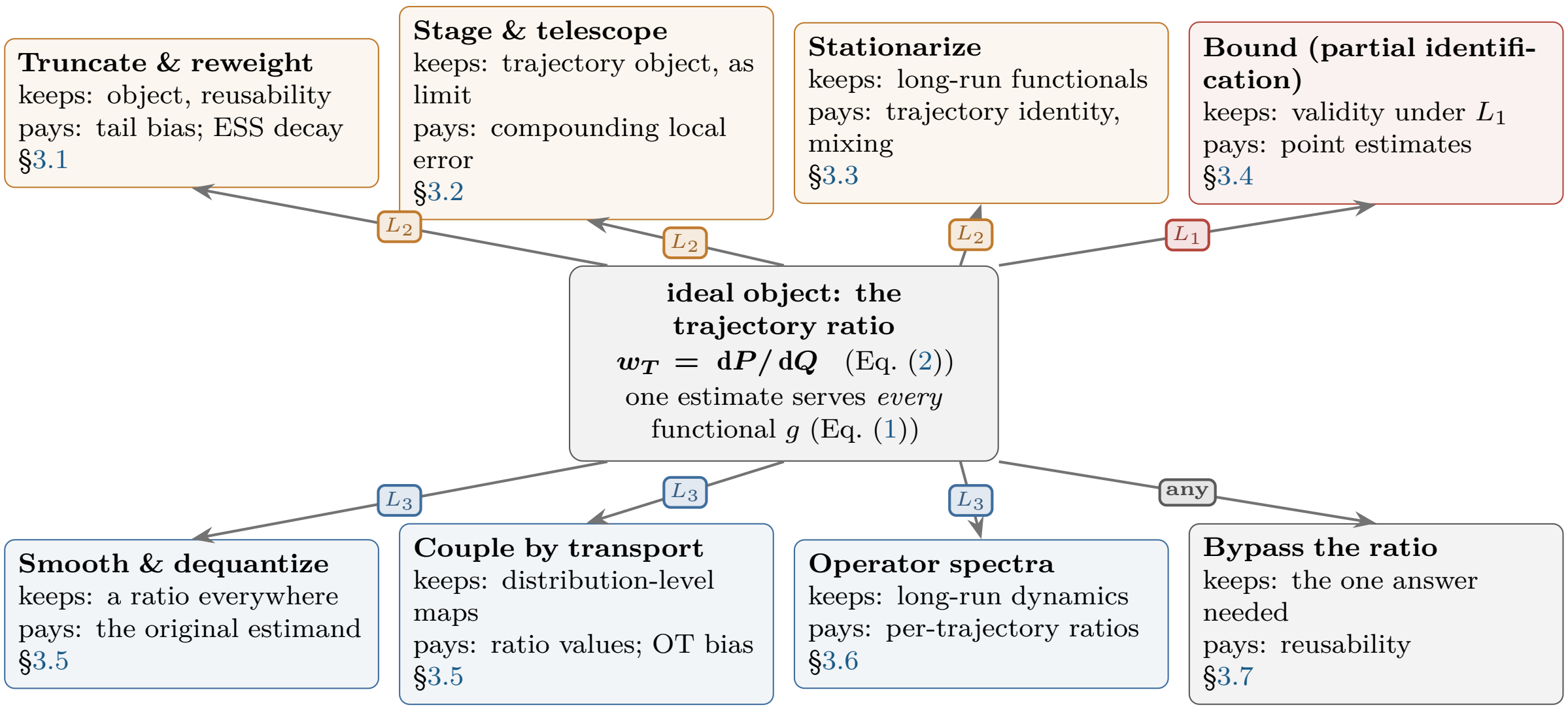
\begin{figure}[tp]
\centering
\resizebox{\linewidth}{!}{\input{figures/fig_retreats}}
\caption{The map of retreats, the organizing claim of this survey. The trajectory ratio
$w_T$ (center) is the only estimand that one estimate reuses across all trajectory functionals
(Eqs.~\eqref{eq:change-of-measure}--\eqref{eq:product}). Each arrow is a retreat to a surrogate
estimand, labeled by the weak-support level that forces it. Each box states what the surrogate
keeps, what it pays, and where it is developed. Table~\ref{tab:retreats} states the same
map on one page with the standardizing domains.}
\label{fig:retreats}
\end{figure}

Read through this definition, the estimator families of Sections~\ref{sec:reweighting}
through~\ref{sec:bypass} are not parallel tribes but nine retreats from one object, ordered by how
much they give up. Direct estimation stays with $w_T$ and pays only while the overlap holds;
truncation and self-normalization keep the object at the price of tail bias. Staging and
telescoping split one ratio into easy ones at the price of compounding local error;
stationarization trades the trajectory for the horizon. Partial identification keeps validity under
$L_1$ at the price of point estimates. Smoothing and dequantization keep a ratio-like object
everywhere at the price of the original estimand. Couplings replace ratios by transport plans;
operator spectra replace densities by spectra, and bypassing the ratio abandons reusability and
estimates one functional directly.
Figure~\ref{fig:retreats} draws the map, and Table~\ref{tab:retreats}
(Appendix~\ref{app:retreats}) states it on one page together with the domains that standardized
each retreat. The remainder of the survey is organized by this map: Section~\ref{sec:tour}
develops the retreats in the order of the map, one subsection each, with the domains that
standardized them folded in as closing passages. Section~\ref{sec:domains} closes the loop from
the domain side, and Section~\ref{sec:open} asks how to retreat later and return sooner.

%% file: figures/fig_retreats.tex
\definecolor{retL1}{HTML}{B4453B}
\definecolor{retL2}{HTML}{C07A2C}
\definecolor{retL3}{HTML}{3F6E9E}
\definecolor{retAny}{HTML}{5A5A5A}
\begin{tikzpicture}[
  font=\scriptsize,
  box/.style={rectangle, rounded corners=3pt, draw=#1, fill=#1!7,
              text width=36mm, minimum height=15mm, align=left, inner sep=4pt},
  badge/.style={fill=#1!15, draw=#1, rounded corners=2pt, font=\tiny\bfseries,
                inner sep=1.8pt, text=#1!80!black},
  center/.style={rectangle, rounded corners=4pt, draw=black!65, fill=gray!10,
                 text width=41mm, minimum height=17mm, align=center, inner sep=5pt},
  arrow/.style={-{Stealth}, black!55, thick}
]

\node[center] (W) at (0,0)
  {\textbf{ideal object: the trajectory ratio}\\[1pt]
   $\bm{w_T=\dd P/\dd Q}$ \; (Eq.~\eqref{eq:product})\\[1pt]
   one estimate serves \emph{every}\\
   functional $g$ (Eq.~\eqref{eq:change-of-measure})};

\node[box=retL2] (r1) at (-6.15, 2.6)
  {\textbf{Truncate \& reweight}\\
   keeps: object, reusability\\
   pays: tail bias; ESS decay\\
   \S\ref{sec:reweighting}};
\node[box=retL2] (r2) at (-2.05, 2.6)
  {\textbf{Stage \& telescope}\\
   keeps: trajectory object, as limit\\
   pays: compounding local error\\
   \S\ref{sec:sequence}};
\node[box=retL2] (r3) at (2.05, 2.6)
  {\textbf{Stationarize}\\
   keeps: long-run functionals\\
   pays: trajectory identity, mixing\\
   \S\ref{sec:dice}};
\node[box=retL1] (r4) at (6.15, 2.6)
  {\textbf{Bound (partial identification)}\\
   keeps: validity under $L_1$\\
   pays: point estimates\\
   \S\ref{sec:partial}};

\node[box=retL3] (r5) at (-6.15,-2.6)
  {\textbf{Smooth \& dequantize}\\
   keeps: a ratio everywhere\\
   pays: the original estimand\\
   \S\ref{sec:bridges}};
\node[box=retL3] (r6) at (-2.05,-2.6)
  {\textbf{Couple by transport}\\
   keeps: distribution-level maps\\
   pays: ratio values; OT bias\\
   \S\ref{sec:bridges}};
\node[box=retL3] (r7) at (2.05,-2.6)
  {\textbf{Operator spectra}\\
   keeps: long-run dynamics\\
   pays: per-trajectory ratios\\
   \S\ref{sec:operators}};
\node[box=retAny] (r8) at (6.15,-2.6)
  {\textbf{Bypass the ratio}\\
   keeps: the one answer needed\\
   pays: reusability\\
   \S\ref{sec:bypass}};

\draw[arrow] ([xshift=4mm]W.north west) -- (r1.south) node[badge=retL2, midway] {$L_2$};
\draw[arrow] (W.north) -- (r2.south) node[badge=retL2, midway] {$L_2$};
\draw[arrow] ([xshift=-4mm]W.north east) -- (r3.south) node[badge=retL2, midway] {$L_2$};
\draw[arrow] (W.north east) -- (r4.south) node[badge=retL1, midway] {$L_1$};
\draw[arrow] ([xshift=4mm]W.south west) -- (r5.north) node[badge=retL3, midway] {$L_3$};
\draw[arrow] (W.south) -- (r6.north) node[badge=retL3, midway] {$L_3$};
\draw[arrow] ([xshift=-4mm]W.south east) -- (r7.north) node[badge=retL3, midway] {$L_3$};
\draw[arrow] (W.south east) -- (r8.north) node[badge=retAny, midway] {any};
\end{tikzpicture}

%% file: sections/survey_r1.tex
\section{Retreats from the trajectory ratio}\label{sec:tour}

This section walks the map one retreat at a time, in the order of
Figure~\ref{fig:retreats} and Table~\ref{tab:retreats}. Each subsection states the surrogate
estimand, develops the machinery, quantifies the price, and closes with the domains that
standardized the retreat. No subsection claims superiority over the others. The choice among them
is the diagnosis problem of Section~\ref{sec:formalization}.

\subsection{Estimating the ratio directly}\label{sec:reweighting}

The first two rows of the map barely retreat at all. Direct estimation learns the ratio function
itself and reuses it across functionals. Reweighting keeps the same object and buys control of its
tails with truncation, self-normalization, and resampling. This section develops the
direct-estimation machinery, the importance-sampling and change-of-measure machinery, and the
domains where they are standardized. Table~\ref{tab:retreats} lists the price.

\subsubsection{Direct density-ratio estimation under strong overlap}\label{sec:kernels}

In the i.i.d.\ setting, density-ratio estimation has mature loss functions. KLIEP directly
minimizes the Kullback--Leibler divergence. LSIF and uLSIF turn ratio estimation into regression by
least squares \citep{sugiyama2007direct,kanamori2009ulsif,sugiyama2012book}. Convex risk
minimization links divergence functionals to likelihood-ratio estimation
\citep{nguyen2010convex,sugiyama2012bregman}, and class-probability estimation unifies the losses
of logistic, probit, and least-squares estimators \citep{menon2016}. Noise-contrastive estimation
fits into the same grid by learning the ratio against a synthetic reference
\citep{gutmann2010,gutmann2012}. The convergence rates of regularized least squares are governed by
the complexity of the function class and the choice of regularization
\citep{caponnetto2007optimal,tsybakov2009introduction,golowich2018size,grunwald2020fast}. When the
ratio is unbounded, variance-penalized and robust estimators restore sub-Gaussian confidence
\citep{maurer2009empirical,catoni2012challenging}, and the nonasymptotic analyses now reaching deep
ratio classes and dependent data extend this line \citep{zheng2026bdd,kengne2026dependent}.

These methods address $L_2$ by regularization and model selection rather than by restoring
identifiability. They behave well inside the support, but the ratio model is unconstrained outside
the data support. When target mass falls in an $L_1$ region, the least-squares objective remains
well defined while returning extrapolation rather than estimation, the regression analogue of
quasi-complete separation in logistic regression \citep{albert1984separation}. Covariate shift is
the canonical application where this failure mode is understood and where the weighting recipe
originated \citep{shimodaira2000improving,sugiyama2007direct,huang2006correcting,bickel2009discriminative}.
distribution-generalization bounds quantify when domain adaptation is possible at all
\citep{bendavid2010theory}, and representation-learning approaches sidestep explicit reweighting
\citep{ganin2016domain}. The density-chasm regime, in which the endpoints are so far apart that
the informative region carries exponentially small mass under both laws, is the same condition observed by
classification-based estimators \citep{srivastava2023,rhodes2020tre}. For trajectory data, time
dependence adds a further caveat: the i.i.d.\ sample-complexity results do not transfer verbatim
to mixing trajectories, and stability bounds for mixing processes set the price of dependence
\citep{mohri2010stability}. Finally, an exponentiated critic score can be trusted as a ratio value,
not merely as a discriminator, only under separate calibration checks
\citep{guo2017calibration,cranmer2015calibrated}.

\subsubsection{Importance sampling and particle filters}

The most direct approach to the trajectory ratio estimates the product in
equation~\eqref{eq:product}. Importance sampling reweights with $w_T$. Sequential Monte Carlo and
particle filters propagate the recursion $w_t=w_{t-1}r_t$ and resample when weights degenerate
\citep{gordon1993novel,doucet2001sequential,tokdar2010importance}. The difficulty is weight
degeneracy: the effective sample size decays rapidly with the horizon and the state dimension, and
in high-dimensional systems particle filters collapse
\citep{bengtsson2008curse,agapiou2017importance}. The classical remedies are variance-reduction
devices known since the beginnings of Monte Carlo, from control of the one-step tails to safe and
bounded proposals and self-normalization
\citep{kahn1953methods,owen2000safe,glynn1989importance,rubinstein2016simulation,owen2013monte},
together with modern intrinsic-dimension analyses showing that the practical cost of importance
sampling is governed by the $\chi^2$-divergence between target and proposal
\citep{agapiou2017importance,chatterjee2018is,bucklew2004introduction}. Equivalent-weights particle
filters push the weights toward the observations to slow degeneracy
\citep{ades2014equivalent,vanleeuwen2009particle}, and ensemble Kalman filters replace full trajectory
weights with a Gaussian approximation of the forecast update
\citep{houtekamer2016review,evensen1994sequential,kalman1960new}, already a half-step toward
the model-substitution retreat of Section~\ref{sec:bypass}.

\subsubsection{The Girsanov change of measure}

Continuous time gives a tighter description. When two diffusions share the diffusion coefficient
$\sigma$ and differ only in the drifts $b_P$ and $b_Q$, the Girsanov theorem writes the trajectory
ratio as the exponential martingale
\begin{equation}\label{eq:girsanov}
  w_T \;=\; \exp\!\left(\int_0^T \sigma^{-1}(b_P-b_Q)\,\dd W_Q
  \;-\; \frac12\int_0^T \bigl\|\sigma^{-1}(b_P-b_Q)\bigr\|^2\dd t\right),
\end{equation}
where $W_Q$ is a Brownian motion under $Q$
\citep{girsanov1960measures,nualart2006malliavin,pavliotis2014stochastic,karatzas1991brownian}.
Equation~\eqref{eq:girsanov} exhibits the trajectory ratio as an exponential martingale: its
expectation is one, but its variance grows with the drift gap and the horizon, the continuous-time
face of the horizon curse, familiar in stochastic thermodynamics as the fluctuation-theorem
martingales of driven processes
\citep{jarzynski1997work,crooks1999fluctuation,seifert2012stochastic}. Malliavin calculus allows
trajectory functionals to be differentiated, connecting sensitivities to reweighting
\citep{fournier1999applications,gobet2005sensitivity,chen2007malliavin}, and continuous-time
reinforcement learning exploits the same martingale structure to write value functions and policy
gradients as martingale representations of trajectory functionals
\citep{jia2022policy,jia2022gradient,wang2020reinforcement}. When the diffusion coefficients differ,
equation~\eqref{eq:girsanov} no longer applies, the trajectory measures are typically mutually singular,
and the problem falls into $L_3$. Noise injection, dequantization, or bridge relaxations are then
unavoidable (Section~\ref{sec:bridges}).

\subsubsection{Instances in the applied sciences}\label{sec:r1-instances}

Three domains run these estimators at production scale, and each has named the trigger in its own language.

\paragraph{Free-energy calculation.}
Zwanzig's perturbation expansion and Kirkwood's coupling-parameter method write the free-energy
difference between two states as an exponential average along a coupling path
\citep{zwanzig1954high,kirkwood1935coupled}, with Bennett's acceptance ratio as the
variance-optimal combination of samples from both ends \citep{bennett1976}. Metropolis--Hastings
sampling supplies the machinery for visiting the joint measure
\citep{metropolis1953equation,hastings1970monte}, and enhanced-sampling methods reshape it
\citep{laio2002escaping,barducci2008well}. Nonequilibrium switching extends reweighting to driven
trajectories \citep{gapsys2020large}. Good-practice guides quantify how far the reweighting can be pushed
before the exponential tails dominate \citep{pohorille2010}. The staging response to the same
pressure is the subject of Section~\ref{sec:staging}.

\paragraph{Monitoring with windows.}
Windowing keeps the direct estimand but localizes it: \citet{kawahara2012changepoint} organize
continuous observations into subsequences for online direct density-ratio change-point detection,
and \citet{liu2013rulsif} compare adjacent windows through relative density ratios and relative
Pearson divergences \citep{yamada2013relative}. The window $\rmZ_t=(\rmX_t,\dots,\rmX_{t+m-1})$ retains
finite-range order so that the static risks serve dynamic detection. Fixed windows, however,
identify only the windowed ratio, and sliding construction adds between-window dependence on top
of within-window dependence. Recursive kernel statistics trade state for compute in the same
monitoring role \citep{keriven2018newma}, and learning-based change detection plugs learned ratios
into cumulative schemes \citep{adiga2022drecusum} while online re-estimation handles drifting
margins \citep{zhang2023online,tibshirani2019weightedcp}. The same streams are what time-series
forecasters model: continuous-time sequence models forecast power-load and sensor series from their
observed histories \citep{li2025odeattention}.

\paragraph{Sim-to-data reweighting and policy lag.}
Particle-physics sim-to-data correction pulls a simulated distribution onto data in a
high-dimensional feature space: boosted decision trees serve as density-ratio models
\citep{rogozhnikov2016reweighting}, neural networks perform full phase-space reweighting and
parameter tuning \citep{andreassen2020neural}, and OmniFold iterates reweighting to unfold all
observables simultaneously \citep{andreassen2020omnifold}. Discriminative decorrelation schemes
modify the same correction through classifier outputs \citep{dolen2016thinking}. In production
reinforcement learning the same retreat appears as truncated importance weights under policy lag
\citep{espeholt2018impala} and as the discrepancy between logged and deployed distributions
\citep{xie2019mis,jiang2020minimax}. In all three cases the failure mode is the reweighting one: when a region
has no source samples, no reweighting can create them, which is the trigger for the bounds of
Section~\ref{sec:partial}.

%% file: sections/survey_r2.tex
\subsection{Splitting the product}\label{sec:sequence}

\begin{figure}[t]
\centering
\resizebox{\linewidth}{!}{\input{figures/fig_timeline}}
\caption{Two research lineages converge on trajectory density-ratio estimation. Top lane: process
measures and weak support, from coupling free energies and sequential analysis to overlap
diagnostics, entropic transport, and dequantified bridges. Bottom lane: sequence-level learning,
from sequential analysis and direct risk minimization to prefix objectives, marginal corrections,
and simulation-calibrated testing. Labels name representative milestones. The map that organizes
both lanes is stated in Table~\ref{tab:retreats}.}
\label{fig:timeline}
\end{figure}
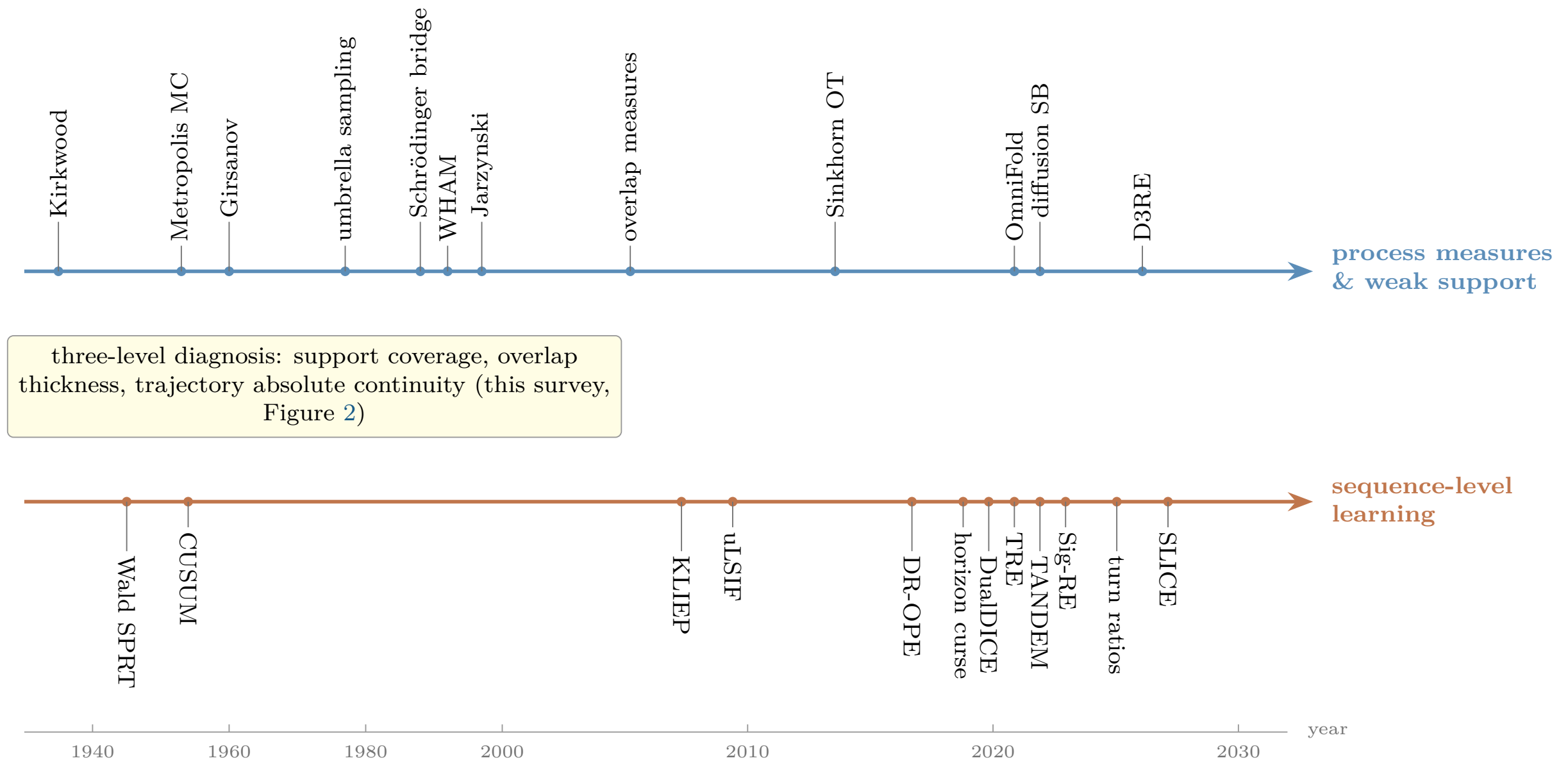

The families of Section~\ref{sec:reweighting} estimate the trajectory ratio as a single object over
the whole trajectory. The second retreat keeps the object but gives up on estimating it in one step: it
splits the product into easy ratios, along a coupling path in configuration space, along a chain of
intermediate distributions, or along the time axis itself. Its levels are characteristically $L_2$
(the product structure amplifies local error) and $L_3$ (the history-conditional view must be
rebuilt when states are hidden), and its key cost is that local estimation error compounds along
the split. This section follows the retreat from its oldest instance, thermodynamic staging, through
distribution-space telescoping to the sequential decompositions that are the working machinery of
simulation-based inference and language-model post-training.
Figure~\ref{fig:timeline} places the milestones of this family and of the weak-support line of
Sections~\ref{sec:reweighting}--\ref{sec:operators} on a common time axis. The two lanes converge
on the trajectory density ratio that this survey is about.

\subsubsection{Staging in free energy and rare events}\label{sec:staging}

The oldest instance of this retreat stages a hopeless ratio into a chain of adjacent ones. Umbrella sampling
partitions configuration space into biased windows \citep{torrie1977nonphysical}, and WHAM merges
the histograms through its self-consistency equations \citep{kumar1992weighted}. In classical
statistics the same idea appears as bridge sampling and path sampling for normalizing constants
\citep{gelman1998path}. The merge is valid only when adjacent windows overlap sufficiently in
phase space, which is $L_2$ in concrete terms. \citet{wu2005phase,wu2005relative} made the premise computable:
phase-space overlap measures detect the bias of free-energy estimates and inform the staged
thermodynamic paths. This practice of running overlap diagnostics before estimation is the most
empirically grounded source of the three-level framework, and bridge and path sampling place the
staged pipeline in a general statistical framework \citep{gelman1998path}.

Nucleation and phase transitions push the staging to its extreme. Classical nucleation theory
describes cluster growth and decay and yields analytic rates
\citep{becker1935kinetische,turnbull1949rate,oxtoby1998nucleation,sear2007nucleation}. The
appearance of a critical nucleus is a rare event whose trajectories almost never cross the barrier
in direct simulation. Forward flux sampling converts rare trajectories into samplable segments by
piecewise biasing \citep{allen2009forward}. Two-step nucleation, prenucleation clusters, and
non-classical pathways show that the target process's reaction coordinates may not be covered by
the data process at all, which is the $L_1$ trigger inside a staged pipeline
\citep{vekilov2005two,gebauer2011prenucleation,erdemir2009nucleation,deyoreo2015crystallization}.
Homogeneous and heterogeneous ice nucleation simulations supply concrete thin-overlap cases:
stacking disorder, metastable ice phases, and surface effects thin the overlap on the order
parameters \citep{lupi2014heterogeneous,lupi2017role,russo2014new,sanz2013homogeneous}, while
heterogeneous nucleation on atmospheric aerosols carries the difficulty into cross-scale model
comparison \citep{murray2012ice,hoose2012heterogeneous,page2006heterogeneous}. Numerical nucleation
rates for NaCl and for Lennard-Jones crystals show how critical sizes and ion-attachment kinetics
determine which regions can be sampled at all
\citep{zimmermann2015nucleation,reintenwolde1996numerical,wolde1997enhancement,auer2001prediction}.
Nonequilibrium work identities give the staged ratio a dynamical form: the Jarzynski equality
relates free-energy differences to exponential averages of work along driven trajectories
\citep{jarzynski1997work}, the Crooks theorem relates forward and reverse trajectory measures through
microscopic reversibility \citep{crooks1999fluctuation,seifert2012stochastic}, and single-molecule
experiments verified these identities and recovered folding free energies from work distributions
\citep{liphardt2002equilibrium,collin2005verification,hummer2001free}.

Alchemical free-energy calculations turn the staging trade-off into engineering practice.
Methodological guidelines and blind-test benchmarks require reporting the overlap of adjacent
windows or $\lambda$ states and decomposing estimation error into sampling and model error
\citep{klimovich2015guidelines,paliwal2011benchmark,cournia2017relative}. Net-charge changes
introduce finite-size electrostatic artifacts that make endpoint differences depend on the
charge-correction scheme \citep{reif2013net,ohlknecht2020correcting}, and small force-field or
binding-site differences reshape the free-energy surface and thereby the overlap region
\citep{rocklin2013calculating}. Two lessons transfer to statistical methodology: overlap
diagnostics can precede estimation, and staging is not a numerical trick but a decomposition of
long $L_1$ transitions into sequences of short $L_2$ transitions, homologous to stationarization
in reinforcement learning (Section~\ref{sec:dice}) and to the bridge constructions of
Section~\ref{sec:bridges}.

\subsubsection{Telescoping in distribution space}\label{sec:telescoping}

A second instance stages the ratio along a chain of \emph{intermediate distributions} rather than
physical coupling parameters. Telescoping density-ratio estimation splits one hard ratio into a
product of adjacent easy ones \citep{rhodes2020tre}. Density-ratio estimation via infinitesimal
classification takes the continuum limit \citep{choi2022}. Multinomial logistic regression
estimates the chasm case directly \citep{srivastava2023}. Conditional probability paths provide
tractable conditional targets with estimation guarantees \citep{yu2025}. Path-variance principles
choose the interpolation to stabilize the resulting estimators \citep{chen2026minpv,chen2026osdre},
and action matching learns the drift directly from samples \citep{neklyudov2023action}. The price
is the same as in thermodynamic staging: the estimand is recovered only through the composed
product, and every intermediate ratio contributes its own estimation error.

\subsubsection{Prefix log-ratio learning}\label{sec:prefix}

The sequential-analysis inheritance supplies the decision-theoretic motivation: Wald's sequential
probability ratio test admits accept, reject, and continue decisions at every step and is optimal
among all tests with the same error probabilities
\citep{wald1945sequential,wald1948optimal}. CUSUM schemes react to a change in distribution with
minimal delay \citep{page1954continuous,lorden1971procedures,basseville1993detection}, and the
sequential-analysis program analyzed stopping rules, boundary crossing, and repeated significance
\citep{siegmund1985sequential,lai2001sequential}. What this inheritance specifies is how to use
evidence arriving step by step. How to \emph{learn} the evidence function itself, namely the prefix log-ratio, from finite samples
of unknown processes is the contribution of the methods below.

SPRT-TANDEM turns the prefix log-ratio itself into the estimation target, jointly optimizing
classification speed and accuracy as evidence accumulates \citep{ebihara2021tandem}. The chain rule
gives
\begin{equation}\label{eq:chain}
  L_T \;=\; \sum_{t=1}^{T}\log\frac{p(\vx_t\mid \vx_{<t})}{q(\vx_t\mid \vx_{<t})},
\end{equation}
and TANDEM forms local increments from source posteriors on adjacent-length segments under a
finite-order Markov approximation. The construction must correct for the history distribution: the
odds of a source classifier on the pair $(\vh,\vx)$ contains the joint ratio $p(\vh,\vx)/q(\vh,\vx)$, not the
conditional ratio $p(\vx\mid \vh)/q(\vx\mid \vh)$, and the identity
\begin{equation}\label{eq:prefix-correction}
  \frac{p(\vx\mid \vh)}{q(\vx\mid \vh)} \;=\; \frac{p(\vh,\vx)/q(\vh,\vx)}{p(\vh)/q(\vh)}
\end{equation}
is what makes local differences accumulate into prefix evidence. The same one-step conditional
factorization is what probabilistic forecasters parameterize: autoregressive models of time series
estimate $p(\vx_t\mid\vx_{<t})$ under a single process, without forming a ratio to a second one
\citep{li2024deepar}. LSEL extends the object to
multi-class density-ratio matrices and proves consistency under finite horizon, common support,
realizability, uniform convergence of the empirical risk, and identifiability
\citep{miyagawa2021lsel}. TANDEMformer analyses then show that internal over-normalization
saturates the log-ratio, sharpening the question from ``can a prefix ratio be built'' to ``how is a
usable ratio magnitude preserved'' \citep{ebihara2023saturation}. Because local estimates add,
$\widehat L_T-L_T=\sum_t(\widehat\ell_t-\ell_t)$, the logarithmic form alone does not guarantee
square-root error growth: lengthwise accumulation and model misspecification require separate
analysis. FIRMBOUND continues the early-classification line with learned stopping boundaries that
are statistically consistent over a finite horizon \citep{ebihara2025firmbound}. None of these
guarantees converts into an arbitrary-stopping error control for a fixed trained network without
the additional arguments of Section~\ref{sec:seq-inference}.

\subsubsection{Ratios without a likelihood}\label{sec:sbi}

When one mechanism can be simulated but not evaluated, the ratio is learned against simulated
reference data. LFIRE performs likelihood-free inference by estimating $p(\vx\mid\theta)/p(\vx)$ with
logistic regression on simulated samples \citep{thomas2022lfire}. DIRE applies convolutional
summaries to dynamic observations \citep{dinev2018dire}, and amortized neural ratio estimation
parameterizes the ratio by the simulator parameters \citep{hermans2020nre}. For a fixed observation
and a common marginal denominator, the quotient of two parameter-conditional ratios connects to the
trajectory likelihood ratio of two mechanisms, the interface through which entire time series enter
simulation-based inference. The field has since built its own evaluation culture: multi-metric
benchmarks \citep{lueckmann2021benchmark}, documented overconfidence of approximate posteriors
\citep{hermans2021trust}, balanced constraints that restore reliability at finite simulation
budgets while preserving the Bayes-optimal solution \citep{delaunoy2022bnre}, bias diagnostics for
contrastive ratio estimators \citep{miller2022cnre}, and simulation-based calibration of posterior
distributions \citep{talts2018validating}. Surveys consolidate the landscape
\citep{cranmer2020frontier}, and the older conditional-density and sequential variants
\citep{papamakarios2016fast,greenberg2019automatic,papamakarios2019sequential,gongalves2020training}
share the same primitive: a learned ratio standing in for an unavailable likelihood. Mining the
simulator for ratios and scores beyond reweighting sharpens the same primitive into grad information
\citep{brehmer2020mining}. The representational caveat remains: summary statistics can compress
away dynamical information, and whole-trajectory classification needs proportionally more data.

Composition replaces whole-trajectory discrimination when transitions can be simulated locally.
Compositional simulation-based inference links local transition simulation at a given state to
global inference through factorized neural ratio estimation, with history-augmentation and
time-conditioning discussed in its appendix \citep{gloeckler2025markovsbi}. Relative to whole-trajectory
representations, the computational organization becomes local learning with length reuse. Relative
to TANDEM, the decisive axis is not whether local composition exists but whether local data can be
generated at a specified state. That difference bounds direct transfer: two mechanisms with
different natural access to histories confound history ratios with conditional ratios in logged
transition classifiers, the original FNRE experiments report local-error sensitivity and
long-sequence degradation, and hidden-state models fall outside its direct scope
\citep{gloeckler2025markovsbi}. For long-memory and non-Markov processes, trawl-process inference
shows parameter telescoping that does not reduce to decomposition along observation time
\citep{leonte2025trawl}.

\subsubsection{Path representations}\label{sec:signatures}

Signatured ratio estimation departs from local multiplication altogether: the whole sequence is
embedded by its path signature, and the ratio is learned globally on that representation
\citep{dyer2022signature}. Its universality statement covers bounded-variation trajectory classes with
fixed starting points and monotone coordinates under parametrized kernels, a function-approximation
guarantee rather than a finite-sample error rate at arbitrary lengths or sampling frequencies. The
broader foundation, namely signature moments as characteristics of the law of a stochastic process,
supports
comparing process distributions through signature features \citep{chevyrev2022signature}. Signature kernels
avoid multiplying explicit local errors but do not remove distribution separation, representation
compression, or importance-weight degeneracy. They put sequence-level density-ratio estimation on
the same footing as kernel two-sample testing \citep{gretton2012mmd}. In the sequence-modeling
idiom of control, trajectory-centric transformers and diffusion planners treat the return-conditioned
sequence itself as the object, sidestepping the ratio but inheriting the same coverage constraints
\citep{chen2021decision,janner2021offline,janner2022planning}.

\subsubsection{Sequence weighting in autoregressive post-training}\label{sec:seq-weighting}

Language-model post-training is currently the largest consumer of sequence-level ratios with
\emph{known} probabilities, and its design choices quantify how much of the trajectory ratio a task
actually needs. Reinforcement learning from human preferences introduced the learned-reward
loop \citep{christiano2017deep,ziegler2019fine,stiennon2020learning}, instruction tuning scaled it
\citep{ouyang2022training}, and PPO stabilized it with clipped surrogates and a
reference-policy KL penalty \citep{schulman2017proximal}. The KL anchor is itself a per-token
log-ratio against the reference process, and the deep-reinforcement-learning precedent for
balancing improvement against trust-region drift is the same trade-off in a different state space
\citep{schulman2015trust,munos2016safe,silver2016mastering,silver2017mastering,silver2018general,mnih2015human,vinyals2019grandmaster}.
Direct preference optimization removes the reward model and optimizes the response-level
log-ratio between policy and reference under the Bradley--Terry reparametrization
\citep{rafailov2023dpo}. The group-relative variant \citep{shao2024deepseekmath} and reasoning-scale
follow-ups \citep{deepseekai2025r1} made response-level ratios the workhorse of verifiable-reward
training, with REINFORCE-style baselines revisited as a lighter alternative
\citep{ahmadian2024back}. Hindsight-privileged RLVR removes the group sampling altogether and
credits tokens on a single rollout from a turn-level confidence estimate, which is itself a
sequence-level distribution \citep{chen2026hindsight}.

Preference optimization is thus an application of the sequence-level ratio in its own right: the
DPO objective optimizes the log-ratio between the trained and the reference process on the same
sequence, and its surrounding machinery is being rebuilt in that vocabulary. The computation graph
can be pruned stochastically for faster training and lower peak memory \citep{wu2026pruning}. The
behavioral criteria guiding preference optimization can be revised by the policy itself
\citep{chen2026rubricopt}. The coupling between the chosen and rejected branches can be
disentangled, so that suppressing the loser no longer suppresses the winner
\citep{chen2026disentangled}. Deployment-oriented variants for mobile and edge acceleration are
consolidated in a dedicated survey \citep{wu2026dpomobile}.

The scale of the estimated object then becomes the design axis. Response-level ratios (DPO) and
length-normalized response ratios (GSPO) \citep{zheng2025gspo} differ from token-level importance
weights in TIS-DPO \citep{liu2025tisdpo}. Prefix visitation corrections differ from single-token
weights in CTPO, whose unbiased change-of-measure analysis covers prefix-measurable token
advantages but not terminal rewards containing behavioral suffixes \citep{zhang2026ctpo}, and MPO
couples cross-turn advantages with the full turn ratio conditional on history, using a
stop-gradient to separate the forward weight of a complete turn from the backward derivation of
token credit \citep{ren2026mpo}. tlm-DRE sets multi-scale weights by per-turn reliability of the
reference policy inside a Bregman training framework \citep{zhao2026tlmdre}. Length normalization
can change optimization dynamics, but it also changes the change-of-measure weights. Imitation-
versus-RLHF formulations \citep{xiao2025dil}, deployment-shift-corrected preferences
\citep{lodkaew2025iwdpo}, and relative density-ratio alignment stabilizing objectives with mixture
denominators \citep{takahashi2026relativealign} make the estimand drift explicit. Hierarchy and
occupancy-measure constraints extend the same object to multi-turn episodes: ArCHer decomposes
return optimization across turn and token levels \citep{zhou2024archer}, and DMPO shapes multi-turn
preference objectives with occupancy-measure constraints and explicit length treatment
\citep{shi2024dmpo}. The same differentiation appears in asynchronous training, where stale-policy
samples require explicit lag control \citep{fu2025areal}, and in the anatomy of off-policy policy
gradient, which compares trajectory importance weights with KL/baseline alternatives within a
restricted family of autoregressive updates, where consistency means preservation of
optimal-policy stationary points, weaker than unbiased gradients \citep{cao2026anatomy}.
Trajectory-level weighted behavioral cloning uses whole-trajectory classification odds with
truncated weights for contaminated demonstrations \citep{pandian2025weightedbc}, twisted sequential
Monte Carlo learns prefix-conditional potentials for sequence-level probabilistic inference
\citep{zhao2024twisted,feng2025tsmc,liu2026multilevel}, and relative density-ratio alignment
stabilizes alignment objectives with mixture denominators, continuing the relative-ratio line
\citep{takahashi2026relativealign}.

Two lessons carry over from the older literature. First, reward-model overoptimization is the
$L_2$ phenomenon in policy space: the optimized proxy drifts from the true objective exactly as
truncated importance weights drift from the ratio \citep{gao2023scaling,moskovitz2024confronting,bai2022constitutional}.
Second, when probabilities are unknown, weighting degrades into classification odds, and the
distinction between known-probability reweighting and learned sequence ratios becomes an
access-condition question rather than a methodological preference
\citep{gloeckler2025markovsbi,pandian2025weightedbc}.

\subsubsection{From estimation to valid inference}\label{sec:seq-inference}

For the exact prefix ratio of a pair of processes, absolutely continuous on every finite horizon,
\begin{equation}\label{eq:martingale}
  \E_Q\bigl[R_t\mid\gF_{t-1}\bigr] = R_{t-1}, \qquad R_0 = 1,
\end{equation}
so the ratio process is a non-negative martingale, connectable through Ville's inequality to
optional-stopping control \citep{ville1939etude,shafer2019game,ramdas2023savi}. Neural estimates do
not automatically satisfy equation~\eqref{eq:martingale}. Neither does that justify the conclusion
that all learned ratios are uncalibratable. The analysis must specify which construction, with
which additional data and assumptions, makes the inference procedure valid. Anytime-valid
inference supplies the interface: test martingales and e-processes keep type-I control at all
stopping times \citep{ramdas2023savi,grunwald2024safe,shafer2021testing,howard2021time,waudby2024estimating},
and betting-based confidence sequences estimate means under the same discipline
\citep{waudby2024estimating,howard2021time}.

Simulation-based calibration closes the loop for learned scores. SLICE selects a positive score
before observing the current round and calibrates against fresh simulations under the null: under
conditional exchangeability, local normalized e-values have conditional expectation one and their
products form a test martingale \citep{cortinovis2026slice}. Anytime-valid simulation-based
hypothesis testing develops the corresponding theory \citep{forre2026anytime}. Score quality affects
power, while validity depends on the calibration construction and the conditional-simulation
access. The appendix discusses extensions to dependent data conditioned on common histories, and
the open questions concern unknown conditional distributions, approximate filtering, and
simulator mismatch \citep{cortinovis2026slice}. Conformal prediction supplies the parallel
uncertainty interface, comprising weighted splits under covariate shift \citep{tibshirani2019weightedcp},
adaptive intervals \citep{gibbs2021aci,xu2021enbpi}, non-exchangeable coverage
\citep{barber2023nonexchangeable}, and the underlying exchangeability framework
\citep{vovk2005algorithmic,lei2014distribution,angelopoulos2023gentle}, and distribution-level
checks catch miscalibrated ratios before deployment \citep{guo2017calibration,hermans2021trust}.
The pattern matches the classical one: estimation accuracy and inference validity are separate
properties, and the second is not delivered free with the first
\citep{wald1945sequential,siegmund1985sequential}.

\begin{table}[t]
\centering\footnotesize
\caption{The three tensions of the sequential literature.}
\label{tab:tensions}
\setlength{\tabcolsep}{2.5pt}
\begin{tabular}{@{}>{\raggedright\arraybackslash}p{30mm} *{8}{c}@{}}
\toprule
Tension &
\rotatebox{50}{\scriptsize direct risk} &
\rotatebox{50}{\scriptsize prefix objectives} &
\rotatebox{50}{\scriptsize signatures} &
\rotatebox{50}{\scriptsize conditional computation} &
\rotatebox{50}{\scriptsize reuse \& DR} &
\rotatebox{50}{\scriptsize occupancy} &
\rotatebox{50}{\scriptsize coverage reweighting} &
\rotatebox{50}{\scriptsize diagnostics \& testing} \\
\midrule
Representation vs.\ accumulated error &
\checkmark & \checkmark & \checkmark & \checkmark & $\times$ & $\times$ & $\times$ & $\times$ \\
\addlinespace
Whole trajectory vs.\ statistical cost &
$\times$ & $\times$ & $\times$ & $\times$ & \checkmark & \checkmark & $\times$ & $\times$ \\
\addlinespace
Estimation vs.\ valid inference &
$\times$ & $\times$ & $\times$ & $\times$ & $\times$ & $\times$ & \checkmark & \checkmark \\
\bottomrule
\end{tabular}
\end{table}

Table~\ref{tab:tensions} summarizes the three tensions that recur across this family. They are not
staged chronologically. Every recent method can be located by which side of each tension it
occupies. The demands behind them are independent and persistent: scientific simulators that cannot
be evaluated, expensive interaction data that must be reused, multi-turn tasks that require process
evaluation, and calibration under deployment shift
\citep{fu2021opebench,yao2024taubench,lodkaew2025iwdpo,cortinovis2026slice}.

%% file: figures/fig_timeline.tex
\definecolor{laneA}{HTML}{5B8DB8}
\definecolor{laneB}{HTML}{C0764C}
\begin{tikzpicture}[font=\scriptsize, >={Stealth}]

  \draw[-{Stealth}, very thick, laneA] (0,1.35) -- (15.1,1.35)
        node[right=2pt, font=\scriptsize\bfseries, laneA, align=left]
        {process measures\\\& weak support};
  \draw[-{Stealth}, very thick, laneB] (0,-1.35) -- (15.1,-1.35)
        node[right=2pt, font=\scriptsize\bfseries, laneB, align=left]
        {sequence-level\\learning};

  \newcommand\milestone[4]{%
    \pgfmathsetmacro\xc{#1<2000 ? (#1-1930)*0.080 : 5.6 + (#1-2000)*0.30}%
    \pgfmathsetmacro\dy{0.30 + 0.28*#3}%
    \ifnum#2>0
      \fill[laneA] (\xc,1.35) circle (1.6pt);
      \draw[black!55] (\xc,1.35) -- (\xc,{1.35+\dy});
      \node[anchor=west, rotate=90, inner sep=1.2pt] at (\xc,{1.35+\dy}) {#4};
    \else
      \fill[laneB] (\xc,-1.35) circle (1.6pt);
      \draw[black!55] (\xc,-1.35) -- (\xc,{-1.35-\dy});
      \node[anchor=west, rotate=-90, inner sep=1.2pt] at (\xc,{-1.35-\dy}) {#4};
    \fi
  }

  \milestone{1935}{1}{1}{Kirkwood}
  \milestone{1953}{1}{0}{Metropolis MC}
  \milestone{1960}{1}{1}{Girsanov}
  \milestone{1977}{1}{0}{umbrella sampling}
  \milestone{1988}{1}{1}{Schr\"odinger bridge}
  \milestone{1992}{1}{0}{WHAM}
  \milestone{1997}{1}{1}{Jarzynski}
  \milestone{2005}{1}{0}{overlap measures}
  \milestone{2013}{1}{1}{Sinkhorn OT}
  \milestone{2020}{1}{0}{OmniFold}
  \milestone{2021}{1}{1}{diffusion SB}
  \milestone{2025}{1}{0}{D3RE}

  \milestone{1945}{-1}{1}{Wald SPRT}
  \milestone{1954}{-1}{0}{CUSUM}
  \milestone{2007}{-1}{1}{KLIEP}
  \milestone{2009}{-1}{0}{uLSIF}
  \milestone{2016}{-1}{1}{DR-OPE}
  \milestone{2018}{-1}{0}{horizon curse}
  \milestone{2019}{-1}{1}{DualDICE}
  \milestone{2020}{-1}{0}{TRE}
  \milestone{2021}{-1}{1}{TANDEM}
  \milestone{2022}{-1}{0}{Sig-RE}
  \milestone{2024}{-1}{1}{turn ratios}
  \milestone{2026}{-1}{0}{SLICE}

  \pgfmathsetmacro\yax{-4.05}
  \draw[black!40] (0,\yax) -- (14.8,\yax);
  \foreach \yr in {1940,1960,1980,2000,2010,2020,2030}{
    \pgfmathsetmacro\xc{\yr<2000 ? (\yr-1930)*0.080 : 5.6 + (\yr-2000)*0.2875}%
    \draw[black!40] (\xc,\yax) -- (\xc,{\yax+0.09});
    \node[font=\tiny, black!55, anchor=north] at (\xc,{\yax-0.02}) {\yr};
  }
  \node[font=\tiny, black!55, anchor=west] at (14.9,\yax) {year};

  \node[draw=black!40, rounded corners=2pt, fill=yellow!12, align=center]
    at (3.4, 0)
    {three-level diagnosis: support coverage, overlap\\
     thickness, trajectory absolute continuity (this survey,\\
     Figure~\ref{fig:retreats})};
\end{tikzpicture}

%% file: sections/survey_r3.tex
\subsection{Stationarization}\label{sec:dice}

The third retreat trades the trajectory for the horizon: stationarization bypasses equation~\eqref{eq:product}. If the process mixes geometrically, the
long-run behavior of the target process is determined by its stationary distribution, and the
stationary ratio $\zeta=\dd\pi/\dd D$ does not grow with the horizon. \citet{liu2018horizon} used
this observation to break the horizon curse in off-policy evaluation. DualDICE removes the
dependence on the behavior distribution through a change of variables and estimates the discounted
stationary ratio directly \citep{nachum2019dualdice}. GenDICE handles average-reward settings via
distribution matching with a normalization constraint \citep{zhang2020gendice}. GradientDICE uses
the Perron--Frobenius argument to eliminate the divergence term and renders the saddle problem
provably convergent under linear parametrization \citep{zhang2020gradientdice}, and BestDICE shows
that these estimators are regularized Lagrangian forms of a single linear program
\citep{yang2020off}.

The linear program is policy evaluation in occupancy-measure form. In the finite-state case,
\begin{equation}\label{eq:occupancy}
  \max_{d\ge0}\;\sum_{\vs,\va} d(\vs,\va)\,r(\vs,\va)
  \quad\text{s.t.}\quad
  \sum_\va d(\vs,\va) = (1-\gamma)\rho_0(\vs) + \gamma \sum_{\vs',\va'} d(\vs',\va')\,P(\vs\mid \vs',\va'),
\end{equation}
whose dual variables are the $Q$-functions. Writing $\zeta=d/dD$ and introducing Lagrange
multipliers yields the prototype saddle objective
\begin{equation}\label{eq:saddle}
  \min_{\zeta\ge0}\;\max_Q\;
  (1-\gamma)\,\E_{\rho_0}\!\bigl[Q(\vs,\pi(\vs))\bigr]
  + \E_D\!\left[\zeta(\vs,\va)\bigl(r(\vs,\va)+\gamma Q(\vs',\pi(\vs'))-Q(\vs,\va)\bigr)\right],
\end{equation}
with normalization $\E_D[\zeta]=1$ \citep{yang2020off,puterman1994markov}. DualDICE, GenDICE, and
GradientDICE differ mainly in how $\zeta$ and $Q$ are regularized, whether redundant constraints
are kept, and whether the final estimate is built from primal or dual variables. OptiDICE applies
the same construction to policy optimization \citep{lee2021optidice}, CoinDICE to confidence
intervals \citep{dai2020coindice}, and AlgaeDICE to policy gradients from arbitrary experience
\citep{nachum2019algaedice}. The lineage descends from the earliest off-policy estimators, which
already recognized that reweighting a trajectory can be replaced by reweighting its visitations
\citep{precup2000eligibility,thomas2016data,dudik2011doubly}. OptiDICE, CoinDICE, AlgaeDICE, and
the treatment of density ratios as first-class assumption classes in online reinforcement learning
\citep{amortila2024harnessing} extend the same saddle structure across tasks.

Marginal importance weights offer a second entry point. Minimax weight and $Q$-function learning
estimates the importance ratio on the state--action distribution without knowing the behavior
policy, while minimax $Q$-learning swaps the roles of weights and value functions. The two combine
into a doubly robust estimator \citep{uehara2020minimax,kallus2020drl}. \citet{jiang2020minimax}
give value intervals under partial coverage. Subsequent work characterizes the finite-sample
behavior of minimax estimation, the role of completeness assumptions, and fast rates
\citep{uehara2021finite,duan2020minimax}, and recasts $Q$-function estimation as nonparametric
instrumental-variable regression with ill-posedness-determined rates \citep{chen2022well}. Under
time-homogeneous Markov dynamics with a single dependent trajectory, the horizon dependence can be
removed under explicit structural assumptions \citep{kallus2022horizon}. The policy-optimization
counterparts connect these saddle problems to trust-region and policy-gradient methods
\citep{kakade2002approximately,schulman2015trust,sutton2000policy,agarwal2021theory,sutton2018reinforcement,szepesvari2010algorithms}.

Stationarization is not a free cure for weak support. It converts trajectory-level $L_2$ into
stationary-level $L_2$: $\zeta$ still must be finite, the normalization constraint must still hold,
and the mixing time determines when the stationary approximation is credible
\citep{meyn2009markov,norris1997markov,kipnis1986convergence}. When the stationary distributions of
the two processes overlap thinly, the solution of equation~\eqref{eq:saddle} still has large
variance, and the bias of truncation and regularization becomes the dominant error. More
fundamentally, stationarization addresses long-run averages only. For finite-horizon hitting
probabilities, first-passage times, or non-stationary processes, the product structure of
equation~\eqref{eq:product} cannot be circumvented entirely.

\subsubsection{Resource selection in ecology}
Resource selection functions are the most literal static instance of the marginal-ratio retreat.
The use--availability design compares locations animals use with locations available in the
environment. The ratio is the selection strength as a density ratio between a use measure and an
availability measure \citep{johnson1980comparison,boyce1999relating,lele2006weighted,manly2002resource}.
Availability, however, is not a statistical support but an ecological modeling choice: which
habitats count as available determines the denominator measure and hence the boundary of $L_1$
\citep{aarts2013quantifying,northrup2013practical}. \citet{fieberg2021how} show that the same
coefficient can carry different meanings under different availability definitions. The methodological
consequence is direct. If availability is defined as all habitats in the study region, $L_1$ is
small but $L_2$ can be severe. If it is defined as habitats actually reachable, $L_1$ shrinks while
the denominator measure becomes harder to estimate. Evaluating selection functions is therefore a
modeling problem co-determined by the estimand and the support definition
\citep{boyce2002evaluating,matthiopoulos2011generalized}. The same lesson appears in the DICE
family in decision-theoretic form: the retreat to marginals is only as good as the measure being marginals
of.

%% file: sections/survey_r4.tex
\subsection{Partial identification}\label{sec:partial}

The fourth retreat keeps validity and gives up point estimates: when $L_1$ dominates, the only
honest estimands are bounds, or values restricted to a policy class.
\citet{sachdeva2020off} analyze contextual bandits with deficient support and compare three
responses: restricting the action space, extrapolating rewards, and restricting the policy class.
\citet{khan2023off} give sharp partial-identification bounds under Lipschitz smoothness.
\citet{mehrabi2024off} handle square-integrable but unbounded ratios with truncated doubly robust
estimators. The causal-inference literature supplies the operational norms: positivity diagnostics
\citep{westreich2010invited,petersen2010diagnosing,cole2008constructing}, propensity-score
adjustment \citep{rosenbaum1983central}, sensitivity analysis, and structural causal models
\citep{pearl2009causality,hernan2016using}. Estimation from truncated samples, in which observations outside a set are simply absent, is the
statistical core of the $L_1$ regime and is possible with
efficient rates only when the truncation geometry is favorable \citep{daskalakis2018efficient}.

The partial-coverage literature of reinforcement learning reaches the same conclusions from the
sampling side. \citet{wang2021statistical} and \citet{foster2022offline} prove that under
insufficient coverage the sample complexity of any algorithm can grow exponentially with the
horizon. Information-theoretic lower bounds and concentrability conditions locate the source of the
impossibility \citep{chen2019information,munos2016safe}. Pessimistic algorithms penalize actions
outside the data support in exchange for provable guarantees
\citep{jin2021pessimism,rashidinejad2021bridging,xie2021bellman,uehara2022pessimistic}, and the
offline reinforcement learning program built on this principle
\citep{levine2020offline,fujimoto2019off,kumar2020conservative,kostrikov2022offline,kumar2019stabilizing,peng2019advantage}
can be read as a catalog of tractable surrogate objectives for the $L_1$ regime. Doubly robust
estimation recovers efficiency when the nuisance models are correct
\citep{jiang2016dr,dudik2011doubly,tang2020doubly}, high-confidence variants attach finite-sample
certificates \citep{thomas2015high}, and counterfactual risk minimization regularizes logged-data
learning by the importance-weight variance itself \citep{swaminathan2015counterfactual}. In
production-scale systems the same tension appears as truncation of importance weights under policy
lag \citep{espeholt2018impala}. For the batch value-iteration ancestor of these guarantees, the
interplay of function approximation and sampling error was quantified early
\citep{ernst2005tree,riedmiller2005neural,antos2008learning}.

The relation of this route to density-ratio estimation deserves emphasis: partial identification is
not the failure of density-ratio estimation but its correct form under $L_1$. Reporting a bound as
a point estimate is the failure. Misreading partial identification as ``the method is not strong
enough'' obstructs the correct judgment of identifiability boundaries.

\subsubsection{Attribution and selection effects}

\emph{Climate attribution.} Climate extremes and event attribution write the density ratio into
the problem definition. The ratio of extreme-event probabilities under factual and counterfactual
climates is a measure ratio on the event set between two climate processes, and attribution must
estimate it with a single realization of the observational record
\citep{stott2004human,otto2017attribution,trenberth2015attribution,shepherd2016common,vanoldenborgh2021pathways}.
This is $L_1$ by construction: the counterfactual climate is never observed, and support
non-coverage is structural rather than a sampling deficiency. The remedies are models, physical
constraints, and reported bounds. Extreme-value statistics supplies the tail language: extremal
indices and return levels rest on regular-variation assumptions
\citep{katz2002statistics,kharin2005estimating,sillmann2013climate}, and large-deviation theory
justifies the bounds \citep{dembo1998large,varadhan1984large,lucarini2016extremes,bucklew2004introduction}.
Rare-event algorithms, including importance splitting, cloning, and genealogical selection, sit between
retreats: they reweight the sampling measure in trajectory space with a staged, splitting structure to estimate
the large-deviation rate functions
\citep{ragone2017computation,ragone2019computation,galfi2019large,bouchet2019rare,touchette2009large}.
The chaotic, high-dimensional character of climate dynamics makes trajectory measures acutely sensitive
to initial conditions, so $L_3$ singularity and low-dimensional structure cannot be avoided either.
The domain thereby combines bounds and model substitution (counterfactuals are simulated by models, not
reweighted (Section~\ref{sec:bypass}).

\emph{Causal inference and process likelihoods.} Causal inference treats positivity as the
identifiability premise and has developed the corresponding diagnostics and sensitivity analyses
\citep{westreich2010invited,petersen2010diagnosing,hernan2016using,rosenbaum1983central,cole2008constructing,pearl2009causality}.
Process models in epidemiology and phylogenetics appear as likelihoods and model comparisons
\citep{fraser2009pandemic,nouvellet2018simple,baele2012accurate,suchard2018bayesian,fan2010choosing}.
They have no explicit density ratio, but the marginal-likelihood ratios between models and the
truncation of the true transmission process by the observation process are governed by the same
support conditions. The common feature is that the process model is explicit, so support
conditions can be partially verified: causal inference can check propensity-score lower bounds,
phylogenetics can compare marginal likelihoods across models. These diagnostics do not remove
$L_1$, but they turn ``not identifiable'' from an abstract warning into an operational reporting
item.

\emph{Selection functions in astrophysics and materials.} Astrophysical weak support comes from
selection effects. Hierarchical Bayesian inference with explicit detection-efficiency models
handles incomplete, noisy catalogs for exoplanet population studies
\citep{foremanmackey2014exoplanet,foremanmackey2013emcee}. Approximate Bayesian computation
refines Kepler occurrence rates \citep{hsu2018improving}, and selection-biased multi-observation
estimation extracts distribution parameters \citep{mandel2019extracting}. The observation process
is a truncated or weighted version of the target process, and ignoring selection mistakes $L_1$
for $L_2$. Writing the selection function explicitly is more reliable than post-hoc reweighting.
Materials design supplies the implicit version: failed experiments and adaptive sampling make the
observed data systematically unrepresentative of the target material space
\citep{raccuglia2016machine,xue2016accelerated,lookman2019active,dunn2020benchmarking,xie2018crystal}.
These fields rarely write down $\dd P/\dd Q$, but the data-generating mechanisms create $L_1$
in exactly the above sense. Bringing them into the framework turns ``selection bias'' from domain
folklore into a testable hypothesis.

%% file: sections/survey_r56.tex
\subsection{Changing the estimand}\label{sec:bridges}

The third strategy accepts that exact reweighting is infeasible and changes the estimand. The
smoothing retreat keeps a ratio-like object but computes it after smoothing or dequantization. The
coupling retreat abandons ratio values and moves mass between the two distributions with a
coupling.

\emph{Smoothing and dequantization.} The Schr\"odinger bridge problem seeks the trajectory measure
with given endpoint marginals that minimizes relative entropy
\citep{follmer1988random,leonard2014survey}. It is isomorphic to stochastic optimal transport and
control \citep{chen2016relation,benamou2000computational,jordan1998variational,otto2001geometry,villani2008optimal,santambrogio2015optimal}.
Diffusion Schr\"odinger bridges turned the construction into a generative tool
\citep{debortoli2021diffusion}, resting on the reverse-time representation of diffusions and the
score-learning machinery
\citep{anderson1982reverse,hyvarinen2005,vincent2011connection,song2019generative,song2021scorebased,ho2020denoising,sohldickstein2015deep,karras2022elucidating,lipman2023flow}.
Diffusion-based generators built on the same reverse-time representation are also used for
long-sequence time-series forecasting \citep{li2026diffinformer}.
For density-ratio estimation, D3RE applies the bridge construction and uses dequantization to
handle disjoint or singular supports \citep{chen2025d3re}, the direct smoothing response to $L_3$: a
ratio-like object exists everywhere, but it is the estimand of the smoothed problem, not the
original $\dd P/\dd Q$.

\emph{Couplings.} Entropy-regularized optimal transport provides the lightweight version: the
Sinkhorn algorithm turns the transport problem into differentiable matrix scaling
\citep{cuturi2013sinkhorn}, with sample complexity and convergence analyzed in detail
\citep{genevay2019sample,mena2019statistical,weed2019sharp,peyre2019computational}. In the
trajectory setting, entropy-regularized transport is not reweighting: it delivers a coupling, not
the trajectory density ratio. Replacing the ratio by a transport cost converts an identifiability
problem into a regularization-bias problem, the central $L_3$ trade-off. The advantage of bridge
and coupling methods is that absolute continuity of the two measures is not required. The price is
that the estimand changes, and the solution is sensitive to the entropy level and to endpoint
estimation error.

\subsubsection{A single-cell instance}
Single-cell trajectory inference faces destructive sampling: each cell is measured once, and time
order must be inferred from expression geometry or splicing kinetics
\citep{trapnell2014dynamics,haghverdi2016diffusion,street2018slingshot,setty2019characterization,qiu2017reversed,cao2019single,saelens2019comparison}.
Pseudotime delivers an ordering, not the time evolution of one cell. The density ratio between two
time points is therefore not directly identifiable. RNA velocity adds directional information from
splicing dynamics \citep{lamanno2018rna,bergen2020generalizing}, and lineage tracing records clonal
relationships through CRISPR scars or barcodes
\citep{kalhor2018developmental,chan2019molecular,mckenna2016whole,spanjaard2018simultaneous,raj2018simultaneous,frieda2016synthetic,livet2007transgenic,kester2018single}.
These technologies provide partially observed processes, not complete trajectory samples.

From the retreat-map viewpoint, single-cell data exhibit $L_1$ and $L_3$ simultaneously: cell
states lie on low-dimensional manifolds, expression matrices are sparse with dropout, and target
fates or perturbation responses may fall outside the observed support. The field's standardized
response is the coupling: neural optimal transport learns perturbation responses
distribution-to-distribution \citep{bunne2023learning,tong2020trajectorynet}, with the transport
plan standing in for an unavailable density ratio, at the price the coupling always pays: no ratio
values, and regularization bias at the marginals. Method-comparison studies repeatedly report sensitivity to
preprocessing, root choice, and topology assumptions
\citep{saelens2019comparison,tritschler2019concepts}. These sensitivities are weak support
manifesting on low-dimensional manifolds. In this domain the ``trajectory'' is often a geometric
reconstruction rather than a sample of a process. Applying density-ratio machinery requires first
making explicit which measure is being estimated.

%% file: sections/survey_r7.tex
\subsection{From densities to spectra}\label{sec:operators}

If the process is stationary, a fourth road avoids trajectory densities altogether. The Perron--Frobenius
operator acts on densities, the Koopman operator on observables, and the two are adjoint. Their
spectral decompositions yield stationary densities, decay rates, and almost-invariant sets
\citep{lasota1994chaos,baladi2000positive,mezic2005spectral,budisic2012applied,koopman1931hamiltonian,klus2016numerical}.
Finite-time coherent sets extend this spectral view to non-autonomous and time-dependent flows
\citep{froyland2005statistically,froyland2013analytic,froyland2015rough,shadden2005definition,aref1984stirring}.
Data-driven approximations, including dynamic mode decomposition and its extended and kernelized
variants, estimate the operator and its spectrum from trajectory ensembles
\citep{schmid2010dynamic,williams2015data,korda2018convergence,klus2016numerical}, and sparse
identification recovers governing equations directly \citep{brunton2016discovering}. Neural ordinary
differential equations learn the governing vector field from trajectories in the same spirit and are
used to forecast continuous-time series \citep{li2024neuralode}. In molecular
dynamics, Markov state models and their variational formulations built the same operator view into
a practical tool for conformational kinetics
\citep{prinz2011markov,nuske2014variational,perezhernandez2013identification,mardt2018vampnets,chodera2014markov,husic2018markov,pande2010everything}.
Boltzmann generators couple such descriptions with learned coordinate changes to sample equilibrium
states of many-body systems \citep{noe2019boltzmann}. For finite Markov chains, the spectral gap of
the transition matrix governs mixing and hence the price of dependence in estimation
\citep{norris1997markov,ethier1986markov,meyn2009markov,jerrum1989approximate}.

These methods address $L_3$ by changing the estimand: rather than the trajectory density ratio, they
estimate an operator or its spectrum. Stationary densities can be read from left eigenvectors, and
ratios between processes approximated through operator perturbation or importance reweighting. The
price is assumptions: spectral gaps, finite-rank approximations, Markovianity, and stationarity.
Perturbation theory and the Fokker--Planck calculus supply the classical instruments for these
approximations \citep{kato1995perturbation,risken1996fokker}. When the process is non-stationary, or
when the weakly supported region corresponds to slow modes with small spectral gaps, operator
estimation error concentrates on those modes and the spectral route fails with it.

%% file: sections/survey_r8.tex
\subsection{Bypassing the ratio}\label{sec:bypass}

The last retreat abandons reusability altogether. When only one comparison or one functional is
needed, the ratio never has to exist as an object: kernel mean embeddings provide another language
in which distributions are embedded in a reproducing kernel Hilbert space and the maximum mean
discrepancy tests whether two distributions differ without estimating densities
\citep{gretton2012mmd,sriperumbudur2010hilbert,muandet2017kernel}, conditional embeddings
characterize transition kernels \citep{fukumizu2007kernel,song2009hilbert}, and nearest-neighbor
divergence estimators bypass the ratio entirely \citep{wang2009divergence,poczos2011nonparametric}.
In the same spirit, generative-adversarial and mutual-information estimators train a critic whose
scalar readout is the answer: the variational bounds are valid objectives, but the impossibility
results for distribution-free mutual-information bounds show they cannot be uniform over all
functionals \citep{goodfellow2014generative,nowozin2016f,arjovsky2017wasserstein,belghazi2018mutual,poole2019variational,oord2018representation,mcallester2020limits}.
Doubly robust value estimation is the decision-theoretic face of the same retreat: the estimand is
the value, and the ratio appears only as a nuisance inside the estimator
\citep{jiang2016dr,dudik2011doubly,tang2020doubly,jiang2020minimax}. The price is uniform across
these variants: the answer does not transfer, because a new functional, threshold, or policy
requires a new estimate and nothing accumulates.

\subsubsection{Instances of model substitution}
The strongest form of this retreat replaces the ratio by a model of the target process itself.
In climate science the counterfactual climate is never observed, so attribution is computed from
model ensembles rather than by reweighting observations, the model-substitution partner of the
bounds of Section~\ref{sec:partial}. In data assimilation the ensemble Kalman filter replaces the
full trajectory weight with a Gaussian approximation of the forecast update
\citep{houtekamer2016review,evensen1994sequential,kalman1960new}. In structural biology,
single-particle reconstruction and structure-prediction systems model the conformational ensemble
statically. Cryo-EM methods infer continuous conformational motions from projections of an
ensemble rather than from trajectories \citep{lyumkis2019challenges,punjani20233dflex}, and
structure-prediction systems such as AlphaFold supply the ensemble statistics directly
\citep{jumper2021highly,senior2020improved}. Endpoint free-energy methods such as MM/PBSA and
MM/GBSA replace path sampling with cheaper approximations and thus provide no path-overlap
diagnostics at all \citep{homeyer2012free,genheden2015mmpbsa}. And in language-model
post-training, the learned reward model of RLHF is precisely a bypass: a scalar proxy stands in
for the return functional, with the familiar bypass price that the proxy must be re-validated for
every new task and drifts under optimization (Section~\ref{sec:seq-weighting}). Time-series
forecasters are the same kind of substitute: a learned model of the series stands in for the process
itself, as in wavelet-based diffusion forecasting \citep{li2025wavelet}.

Model substitution converts the sampling problem into a model-correctness problem: the error is no
longer variance under $Q$-sampling but discrepancy between the substituted model and the target
process. That error is not governed by the overlap diagnostics of direct estimation and reweighting at all, which is why
model-substitution pipelines require their own validation culture of simulation-based calibration,
held-out physics, and benchmark coverage rather than overlap reporting
\citep{talts2018validating,lueckmann2021benchmark}.

%% file: sections/survey_domains.tex
\section{Cross-domain synthesis}\label{sec:domains}

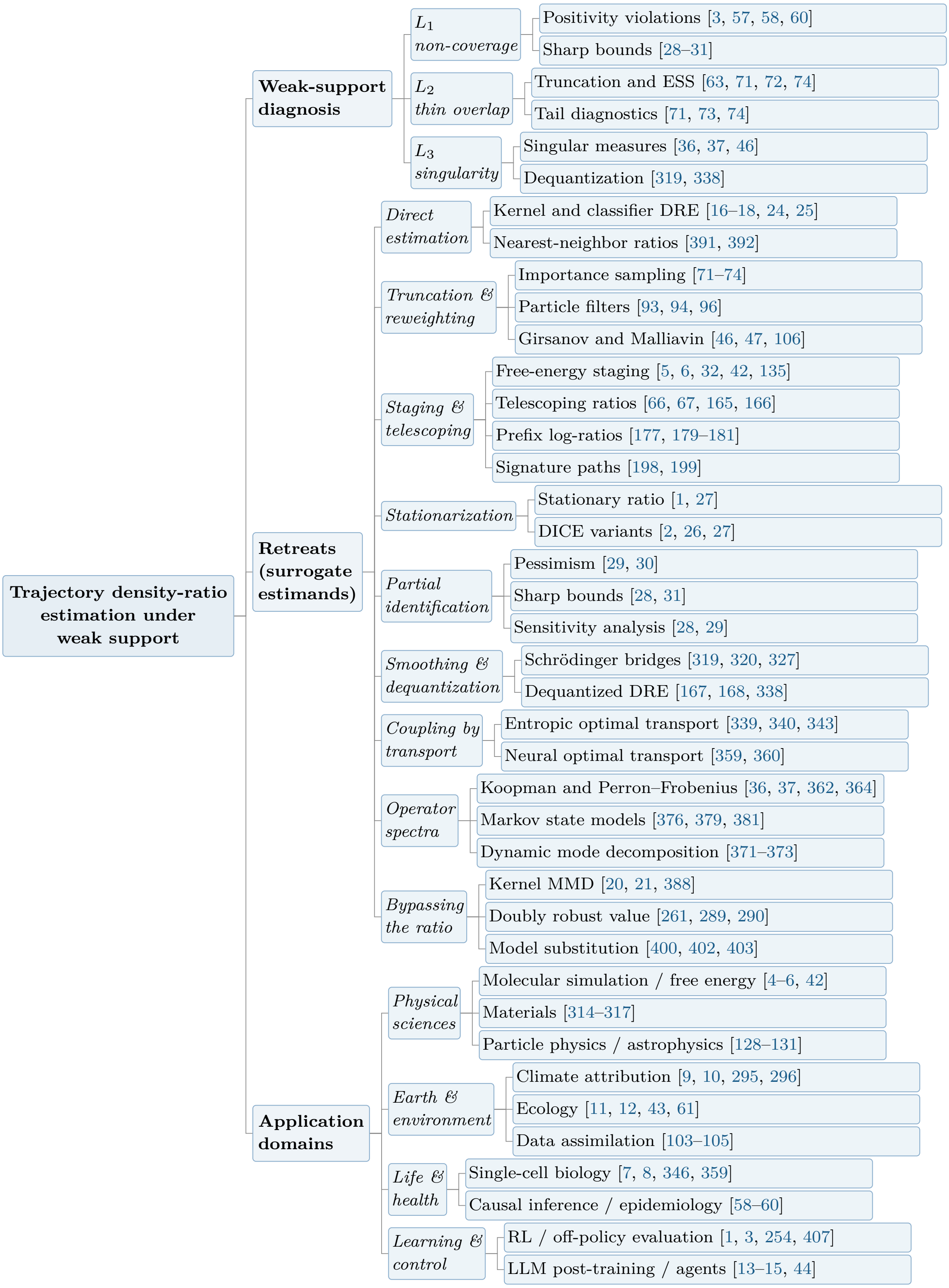
\begin{figure}[p]
\vspace*{\fill}
\centering
\input{figures/fig_taxonomy}
\caption{Taxonomy of the surveyed literature by weak-support diagnosis, retreat, and application
domain. Each leaf lists representative works; a work appears under every category it serves.}
\label{fig:taxonomy}
\vspace*{\fill}
\end{figure}

Section~\ref{sec:tour} developed the retreats one at a time and, at each step, named the domains
that standardized them. This section closes the loop from the other side: first the estimator
families are juxtaposed (Table~\ref{tab:methods}). Then the cross-domain comparison is keyed by
retreat rather than by discipline (Table~\ref{tab:domains}), and finally the newest domain,
language-model post-training and multi-turn agents, is read as a composite in which nearly every
retreat operates at once. Figure~\ref{fig:taxonomy} classifies the corpus along the same three
axes: the weak-support diagnosis, the retreat, and the application domain.

\subsection{The estimator families, juxtaposed}\label{sec:comparison}

Table~\ref{tab:methods} juxtaposes the families as a tick matrix: the three level columns record
which weak-support level each family addresses, and the last columns give the key cost and
representative work. One observation follows immediately: each family covers only one or two
levels. Path reweighting has uncontrollable variance under $L_2$ and fails under $L_3$.
Stationarization converts the horizon amplification of $L_2$ into a mixing-time and
stationary-ratio problem. Direct density-ratio estimation relies on regularization inside the
support. Partial identification changes the estimand. Bridge methods substitute a coupling for the
ratio. Operator methods substitute a spectrum for a density, and sequential decompositions buy
tractability with compounding local error. Deployed systems are typically composites: an operator
or physical model supplies structure, partial identification bounds the $L_1$ regions, and
truncation or regularization controls $L_2$ variance. The open problems of Section~\ref{sec:open}
are exactly the seams where these composites meet. Table~\ref{tab:retreats} in
Appendix~\ref{app:retreats} states the same map with the standardizing domains and key work.

\begin{table}[t]
\centering\footnotesize
\caption{Estimator families by weak-support level
(\checkmark\ = yes, $\circ$ = partly, $\times$ = no).}\label{tab:methods}
\setlength{\tabcolsep}{2pt}
\begin{tabular}{@{}>{\raggedright\arraybackslash}p{24mm}>{\raggedright\arraybackslash}p{19mm}ccc c >{\raggedright\arraybackslash}p{22mm} >{\raggedright\arraybackslash}p{30mm}@{}}
\toprule
Family & Estimand & $L_1$ & $L_2$ & $L_3$ & Horizon-free & Key cost & Representative work \\
\midrule
Path reweighting and particle filters &
trajectory ratio $w_T$ &
-- & \checkmark & $\times$ & $\times$ &
weight degeneracy &
\citet{gordon1993novel}, \citet{bengtsson2008curse} \\
\addlinespace
Girsanov and Malliavin calculus &
continuous-time ratio &
-- & \checkmark & $\circ$ & $\times$ &
common diffusion coefficient &
\citet{girsanov1960measures}, \citet{nualart2006malliavin} \\
\addlinespace
Direct density-ratio estimation &
ratio function &
-- & \checkmark & $\times$ & $\times$ &
no guarantees outside the support &
\citet{sugiyama2007direct}, \citet{kanamori2009ulsif} \\
\addlinespace
Stationarization (DICE family) &
stationary ratio $\zeta$ &
-- & \checkmark & $\times$ & \checkmark &
mixing and normalization &
\citet{liu2018horizon}, \citet{nachum2019dualdice} \\
\addlinespace
Partial identification and pessimism &
bounds &
\checkmark & $\circ$ & $\times$ & \checkmark &
bounds widen with the gap &
\citet{khan2023off}, \citet{jin2021pessimism} \\
\addlinespace
Bridges and entropic optimal transport &
coupling or bridge &
$\circ$ & $\times$ & \checkmark & \checkmark &
the estimand changes &
\citet{follmer1988random}, \citet{debortoli2021diffusion} \\
\addlinespace
Transfer operators and spectral methods &
operator and spectrum &
-- & $\times$ & \checkmark & \checkmark &
spectral gap and stationarity &
\citet{lasota1994chaos}, \citet{mardt2018vampnets} \\
\addlinespace
Sequential decomposition &
prefix log-ratios &
-- & \checkmark & $\circ$ & $\times$ &
compounding local error &
\citet{ebihara2021tandem}, \citet{dyer2022signature} \\
\bottomrule
\end{tabular}
\end{table}

\subsection{The cross-domain comparison, keyed by retreat}\label{sec:cross-domain}

Table~\ref{tab:domains} keys the cross-domain comparison by retreat: a check mark marks the
domains that standardize each retreat in everyday practice, and the last column names the trigger
as each field calls it locally.

\begin{table}[t]
\centering\footnotesize
\caption{Cross-domain comparison keyed by retreat.}\label{tab:domains}
\renewcommand{\arraystretch}{1.3}
\setlength{\tabcolsep}{2pt}
\begin{tabular}{@{}>{\raggedright\arraybackslash}p{23mm} *{9}{c} >{\raggedright\arraybackslash}p{27mm}@{}}
\toprule
Retreat &
\rotatebox{50}{\scriptsize free energy} &
\rotatebox{50}{\scriptsize single cell} &
\rotatebox{50}{\scriptsize climate} &
\rotatebox{50}{\scriptsize ecology} &
\rotatebox{50}{\scriptsize particle \& astro} &
\rotatebox{50}{\scriptsize data assimilation} &
\rotatebox{50}{\scriptsize causal inference} &
\rotatebox{50}{\scriptsize materials} &
\rotatebox{50}{\scriptsize agents} &
Trigger, as named locally \\
\midrule
Direct estimation and reweighting (\S\ref{sec:reweighting}) &
\checkmark & $\times$ & $\times$ & $\times$ & \checkmark & \checkmark & \checkmark & $\times$ & \checkmark &
phase-space overlap, weight degeneracy, policy lag \\
\addlinespace
Staging and telescoping (\S\ref{sec:sequence}) &
\checkmark & $\times$ & $\times$ & $\times$ & $\times$ & $\times$ & $\times$ & $\times$ & \checkmark &
window overlap, barrier rarity \\
\addlinespace
Stationarization (\S\ref{sec:dice}) &
$\times$ & $\times$ & $\times$ & \checkmark & $\times$ & $\times$ & $\times$ & $\times$ & \checkmark &
availability, concentrability \\
\addlinespace
Bounds (partial identification) (\S\ref{sec:partial}) &
$\times$ & $\times$ & \checkmark & $\times$ & \checkmark & $\times$ & \checkmark & \checkmark & \checkmark &
positivity, detection probability \\
\addlinespace
Smoothing and dequantization (\S\ref{sec:bridges}) &
$\times$ & $\times$ & $\times$ & $\times$ & $\times$ & $\times$ & $\times$ & $\times$ & $\times$ &
disjoint or singular supports \\
\addlinespace
Coupling by transport (\S\ref{sec:bridges}) &
$\times$ & \checkmark & $\times$ & $\times$ & $\times$ & $\times$ & $\times$ & $\times$ & $\times$ &
destructive sampling \\
\addlinespace
From densities to spectra (\S\ref{sec:operators}) &
\checkmark & $\times$ & \checkmark & $\times$ & $\times$ & $\times$ & $\times$ & $\times$ & $\times$ &
stationarity, slow modes \\
\addlinespace
Bypassing the ratio (\S\ref{sec:bypass}) &
$\times$ & $\times$ & \checkmark & $\times$ & $\times$ & \checkmark & $\times$ & \checkmark & \checkmark &
unobservable target process \\
\bottomrule
\end{tabular}
\end{table}

Three regularities emerge. First, the support is defined by the data-collection design, not by the
target distribution: availability, detection probability, behavior policy, biasing potential, and
logged-policy support all rewrite the denominator. Second, the diagnostics of different fields are
mathematically homologous: phase-space overlap, effective sample size, positivity, and the overlap
coefficient all measure how close the problem sits to the level that forces a retreat. Third, the
remedies trade among three costs: bias from truncation, regularization, and entropy; identifiability
from partial identification and restricted policy classes; and computation from bridge constructions
and operator approximations. These three regularities are the starting points of the open problems
in Section~\ref{sec:open}.

\subsection{Language-model post-training and multi-turn agents}\label{sec:agents}

The newest domain is also the densest: language-model post-training and its agent deployments
operate on process-generated trajectories by construction, since an episode is a trajectory of tokens,
tool calls, and observations. The data process is the logged policy, and the target process is the
trained or deployed policy. The deep-reinforcement-learning precedent demonstrated that
self-generated trajectories can be converted into superhuman process control when the objective is
verifiable
\citep{mnih2015human,silver2016mastering,silver2017mastering,silver2018general,vinyals2019grandmaster,fawzi2022discovering,degrave2022magnetic}.
Language models inherit the machinery with a stochastic, partially observed environment in the loop
\citep{bommasani2021opportunities,touvron2023llama,openai2023gpt4}. The demand side is dominated by
data reuse: off-policy evaluation for logged interactions asks exactly the trajectory-ratio
question, with the corresponding benchmark and certificate culture
\citep{thomas2015high,fu2021opebench,voloshin2021empirical,espeholt2018impala}. Multi-turn agent
benchmarks make the process character explicit: user--tool--agent interaction with repeated
reliability requirements \citep{yao2024taubench,barres2025tau2}, open-ended web and computer tasks
\citep{zhou2024webarena,yao2022webshop,deng2023mind2web,koh2024osworld}, real software repositories
\citep{jimenez2024swe}, and progress-style metrics beyond terminal success
\citep{ma2024agentboard}. The interacting policies reason and act over long horizons
\citep{yao2023react,nakano2021webgpt,schick2023toolformer}, and intermediate-process supervision
targets the sparse-reward structure of such episodes \citep{lightman2024verify,feng2025gigpo}.

Weak support appears at all three levels, with level-specific names, and the domain therefore
composes nearly every retreat. At the first level, the logged policy assigns zero probability to
actions the deployed system takes, a condition known as deficient support in the logs. User and environment mechanisms change
between training and deployment, so the denominator measure itself moves, and prompts or tool
states arise that the logs never contain \citep{sachdeva2020off,lodkaew2025iwdpo,yao2024taubench}.
At the second level, sequence-level importance weights degenerate over long responses and
multi-turn episodes, which is why the field re-levels the ratio, whether per-response, per-token, per-turn, or
per-prefix, as quantified by the sequence-weighting designs of Section~\ref{sec:seq-weighting}
\citep{schulman2017proximal,shao2024deepseekmath,zheng2025gspo,liu2025tisdpo,zhang2026ctpo,ren2026mpo,zhao2026tlmdre}.
At the third level, the user's intent state is never observed, and the joint user--agent process is
only partially measurable from transcripts. Prefix-conditional potentials and surrogate processes stand
in for the unobservable one \citep{zhao2024twisted,feng2025tsmc}. Reward-model overoptimization is
the $L_2$ drift of the optimized proxy away from the true objective
\citep{gao2023scaling,moskovitz2024confronting}, and the reference-policy KL anchor is a
deliberately induced $Q$ whose overlap with the target policy is kept thick, an explicit overlap
management in the sense of Section~\ref{sec:dice}.

For this domain the map gives a design rule and a research agenda. The rule: when probabilities are
known, choose the weighting level from the estimand (response, turn, or prefix) and control the
tails. When they are unknown, the problem is sequence-level density-ratio estimation under
restricted access, and the methods of Section~\ref{sec:sequence} apply with their compounding-error
caveats \citep{gloeckler2025markovsbi,ebihara2021tandem,pandian2025weightedbc}. The agenda:
multi-turn partial identification, namely what remains identifiable about a turn-level ratio when
the user process is unknown, and horizon-aware rates for trajectory ratios in autoregressive
processes, both of which Section~\ref{sec:open} states precisely.

%% file: figures/fig_taxonomy.tex
\definecolor{taxline}{HTML}{5B8DB8}
\definecolor{taxfill}{HTML}{EDF4F8}
\begin{forest}
  for tree={
    grow'=east, forked edges,
    draw=taxline!70, fill=taxfill, rounded corners=1.5pt,
    font=\scriptsize, inner sep=1.5pt, align=left,
    anchor=west, parent anchor=east, child anchor=west,
    edge={draw=black!45, line width=0.35pt},
    l sep=3mm, s sep=0.35mm,
  },
  where level=0{fill=taxline!15, font=\scriptsize\bfseries, align=center, inner sep=3pt}{},
  where level=1{fill=taxline!10, font=\scriptsize\bfseries, inner sep=2.5pt}{},
  where level=2{font=\scriptsize\itshape, inner sep=1.5pt}{},
  [Trajectory density-ratio\\estimation under\\weak support
    [Weak-support\\diagnosis
      [$L_1$\\non-coverage
        [{\parbox[t]{66mm}{\raggedright Positivity violations \citep{sachdeva2020off,westreich2010invited,thomas2015high,cole2008constructing}}}]
        [{\parbox[t]{66mm}{\raggedright Sharp bounds \citep{khan2023off,jin2021pessimism,wang2021statistical,foster2022offline}}}]
      ]
      [$L_2$\\thin overlap
        [{\parbox[t]{66mm}{\raggedright Truncation and ESS \citep{kahn1953methods,mehrabi2024off,agapiou2017importance,chatterjee2018is}}}]
        [{\parbox[t]{66mm}{\raggedright Tail diagnostics \citep{agapiou2017importance,chatterjee2018is,glynn1989importance}}}]
      ]
      [$L_3$\\singularity
        [{\parbox[t]{66mm}{\raggedright Singular measures \citep{girsanov1960measures,lasota1994chaos,baladi2000positive}}}]
        [{\parbox[t]{66mm}{\raggedright Dequantization \citep{chen2025d3re,follmer1988random}}}]
      ]
    ]
    [Retreats\\(surrogate\\estimands)
      [Direct\\estimation
        [{\parbox[t]{66mm}{\raggedright Kernel and classifier DRE \citep{sugiyama2007direct,kanamori2009ulsif,nguyen2010convex,menon2016,gutmann2012}}}]
        [{\parbox[t]{66mm}{\raggedright Nearest-neighbor ratios \citep{wang2009divergence,poczos2011nonparametric}}}]
      ]
      [Truncation \&\\reweighting
        [{\parbox[t]{66mm}{\raggedright Importance sampling \citep{kahn1953methods,glynn1989importance,agapiou2017importance,chatterjee2018is}}}]
        [{\parbox[t]{66mm}{\raggedright Particle filters \citep{gordon1993novel,bengtsson2008curse,doucet2001sequential}}}]
        [{\parbox[t]{66mm}{\raggedright Girsanov and Malliavin \citep{girsanov1960measures,nualart2006malliavin,karatzas1991brownian}}}]
      ]
      [Staging \&\\telescoping
        [{\parbox[t]{66mm}{\raggedright Free-energy staging \citep{kumar1992weighted,torrie1977nonphysical,bennett1976,gelman1998path,wu2005phase}}}]
        [{\parbox[t]{66mm}{\raggedright Telescoping ratios \citep{rhodes2020tre,srivastava2023,choi2022,yu2025}}}]
        [{\parbox[t]{66mm}{\raggedright Prefix log-ratios \citep{ebihara2021tandem,miyagawa2021lsel,ebihara2023saturation,ebihara2025firmbound}}}]
        [{\parbox[t]{66mm}{\raggedright Signature paths \citep{dyer2022signature,chevyrev2022signature}}}]
      ]
      [Stationarization
        [{\parbox[t]{66mm}{\raggedright Stationary ratio \citep{liu2018horizon,yang2020off}}}]
        [{\parbox[t]{66mm}{\raggedright DICE variants \citep{nachum2019dualdice,zhang2020gendice,yang2020off}}}]
      ]
      [Partial\\identification
        [{\parbox[t]{66mm}{\raggedright Pessimism \citep{jin2021pessimism,foster2022offline}}}]
        [{\parbox[t]{66mm}{\raggedright Sharp bounds \citep{khan2023off,wang2021statistical}}}]
        [{\parbox[t]{66mm}{\raggedright Sensitivity analysis \citep{foster2022offline,wang2021statistical}}}]
      ]
      [Smoothing \&\\dequantization
        [{\parbox[t]{66mm}{\raggedright Schr\"odinger bridges \citep{debortoli2021diffusion,leonard2014survey,follmer1988random}}}]
        [{\parbox[t]{66mm}{\raggedright Dequantized DRE \citep{chen2025d3re,chen2026minpv,chen2026osdre}}}]
      ]
      [Coupling by\\transport
        [{\parbox[t]{66mm}{\raggedright Entropic optimal transport \citep{cuturi2013sinkhorn,peyre2019computational,genevay2019sample}}}]
        [{\parbox[t]{66mm}{\raggedright Neural optimal transport \citep{bunne2023learning,tong2020trajectorynet}}}]
      ]
      [Operator\\spectra
        [{\parbox[t]{66mm}{\raggedright Koopman and Perron--Frobenius \citep{koopman1931hamiltonian,lasota1994chaos,mezic2005spectral,baladi2000positive}}}]
        [{\parbox[t]{66mm}{\raggedright Markov state models \citep{mardt2018vampnets,prinz2011markov,husic2018markov}}}]
        [{\parbox[t]{66mm}{\raggedright Dynamic mode decomposition \citep{schmid2010dynamic,williams2015data,korda2018convergence}}}]
      ]
      [Bypassing\\the ratio
        [{\parbox[t]{66mm}{\raggedright Kernel MMD \citep{gretton2012mmd,muandet2017kernel,sriperumbudur2010hilbert}}}]
        [{\parbox[t]{66mm}{\raggedright Doubly robust value \citep{jiang2016dr,dudik2011doubly,tang2020doubly}}}]
        [{\parbox[t]{66mm}{\raggedright Model substitution \citep{jumper2021highly,senior2020improved,lyumkis2019challenges}}}]
      ]
    ]
    [Application\\domains
      [Physical\\sciences
        [{\parbox[t]{66mm}{\raggedright Molecular simulation / free energy \citep{kirkwood1935coupled,kumar1992weighted,bennett1976,torrie1977nonphysical}}}]
        [{\parbox[t]{66mm}{\raggedright Materials \citep{raccuglia2016machine,lookman2019active,xue2016accelerated,dunn2020benchmarking}}}]
        [{\parbox[t]{66mm}{\raggedright Particle physics / astrophysics \citep{rogozhnikov2016reweighting,andreassen2020omnifold,andreassen2020neural,dolen2016thinking}}}]
      ]
      [Earth \&\\environment
        [{\parbox[t]{66mm}{\raggedright Climate attribution \citep{stott2004human,otto2017attribution,trenberth2015attribution,shepherd2016common}}}]
        [{\parbox[t]{66mm}{\raggedright Ecology \citep{johnson1980comparison,lele2006weighted,aarts2013quantifying,northrup2013practical}}}]
        [{\parbox[t]{66mm}{\raggedright Data assimilation \citep{houtekamer2016review,evensen1994sequential,kalman1960new}}}]
      ]
      [Life \&\\health
        [{\parbox[t]{66mm}{\raggedright Single-cell biology \citep{trapnell2014dynamics,bunne2023learning,saelens2019comparison,setty2019characterization}}}]
        [{\parbox[t]{66mm}{\raggedright Causal inference / epidemiology \citep{westreich2010invited,petersen2010diagnosing,cole2008constructing}}}]
      ]
      [Learning \&\\control
        [{\parbox[t]{66mm}{\raggedright RL / off-policy evaluation \citep{thomas2015high,liu2018horizon,voloshin2021empirical,fu2021opebench}}}]
        [{\parbox[t]{66mm}{\raggedright LLM post-training / agents \citep{ouyang2022training,rafailov2023dpo,yao2024taubench,schulman2017proximal}}}]
      ]
    ]
  ]
\end{forest}

%% file: sections/survey_open.tex
\section{Open problems}\label{sec:open}

In the vocabulary of the retreat map, the open problems are not a scatter of isolated technical
gaps: they ask either to \emph{retreat later}, by diagnosing the trigger before paying for it, to
\emph{retreat less far}, with surrogates that keep more of the trajectory object, or to
\emph{return}, with rates that say when the surrogate can be trusted as a stand-in for $w_T$. We state seven, each with the
evidence that makes it well posed.

\paragraph{1. Diagnostics matched to the three levels.}
$L_1$ can be probed by propensity-score lower bounds and support-coverage tests
\citep{sachdeva2020off,westreich2010invited}; $L_2$ by effective sample sizes, overlap
coefficients, and $\chi^2$ estimation \citep{wu2005phase,agapiou2017importance,gretton2012mmd};
$L_3$ requires trajectory-level singularity probes, for instance the stability of an estimated ratio
under increasing noise scales. Each quantity is practiced somewhere, but there is no common
reporting standard and no theory connecting them to an error decomposition of the estimator. A
useful target is a decomposition of estimation error into an identifiability gap, an
overlap-induced variance, and a function-approximation error, each with an estimable diagnostic.
the impossibility results for distribution-free mutual-information bounds show that such
decompositions cannot be uniform over all functionals and must be stated per object
\citep{mcallester2020limits,poole2019variational}. Such a decomposition would ground method
selection and make cross-domain comparisons commensurable.

\paragraph{2. Path-level partial identification.}
\citet{khan2023off} give sharp bounds and minimax rates in static or single-step settings.
\citet{mehrabi2024off} handle square-integrable weak overlap in MDPs. But when a transition kernel
is missing on a region and the target functional depends on the joint distribution over many
trajectories, sharp bounds for trajectory functionals under mixing, Lipschitz, and Markov structure remain open.
Observation~\ref{obs:escalation} shows that local gaps escalate to trajectory singularity, so trajectory-level
partial identification cannot be obtained by multiplying one-step bounds. New tools are needed to
characterize \emph{which trajectories remain identifiable}
\citep{wang2021statistical,foster2022offline,jin2021pessimism}.

\paragraph{3. A rate theory of time-amplified error.}
Equation~\eqref{eq:product} and Observation~\ref{obs:amplification} give the exponential mechanism,
and stationarization avoids it under mixing, but general conditions determining \emph{when}
stationarization removes the exponential horizon dependence are missing. A joint characterization
of mixing time, overlap thickness, and function-class complexity is the key to understanding when
long trajectories are a blessing and when they are a curse
\citep{chen2019information,meyn2009markov,lezaud1998chernoff,mohri2010stability}. Existing lower
bounds are given on specific MDP classes \citep{wang2021statistical,foster2022offline}. Extensions
to diffusions and continuous-time processes are incomplete, although the martingale structure of
Section~\ref{sec:reweighting} suggests the Girsanov variance $\int_0^T\|\Delta b\|^2\dd t$ is the
right continuous-time proxy for the exponential constant
\citep{girsanov1960measures,jia2022policy}.

\paragraph{4. The estimation object under smoothing.}
Noise injection, dequantization, entropy regularization, and Schr\"odinger bridges convert singular
problems into solvable ones, but they change the estimand: what is recovered is a smoothed ratio or
a coupling, not the original $\dd P/\dd Q$ \citep{chen2025d3re,debortoli2021diffusion,cuturi2013sinkhorn,bunne2023learning}.
How large the induced bias is, under which conditions it is negligible, and how the smoothing scale
should be selected from data are currently empirical questions. Bridges and dequantified variants
are the most promising starting points \citep{chen2025d3re,yu2025}, but the theory lags the
algorithms.

\paragraph{5. Structured extrapolation.}
Existing $L_1$ remedies mostly rely on black-box smoothness (Lipschitz, H\"older) or monotonicity.
process data carry stronger structure: transfer operators, causal graphs, conservation laws, and
physical constraints. Encoding this structure as the assumption class of the ratio function could
yield tighter bounds than generic smoothness. Transfer operators and spectral methods already
estimate stationary densities \citep{froyland2015rough,mardt2018vampnets,brunton2016discovering},
and causal structure constrains which ratios are identifiable at all \citep{pearl2009causality}.
how to use them for non-stationary, weakly supported trajectory functionals is open. Physics-based ratio
estimation points in the same direction: entropy-production estimators exploit time-reversal
structure \citep{kim2020neep,aguilera2026maxent}, diffusion forward--backward trajectory ratios connect
densities, control, and energy training \citep{he2026rne,chen2026minpv}, and nonasymptotic analyses
for deep ratio classes and dependent data are beginning to appear \citep{zheng2026bdd,kengne2026dependent}.
The modeling of density ratios as first-class assumption classes in online reinforcement learning
\citep{amortila2024harnessing} and the use of outcome-augmented likelihood changes for retrospective
action credit \citep{chakrabarti2026relace} suggest that the assumption class itself can be an
algorithm-design object.

\paragraph{6. Optimization and benchmarks.}
DICE-type saddle problems can lose convex--concave structure under nonlinear parametrization
\citep{zhang2020gradientdice}, regularization and truncation introduce bias, and there is no
standard practice for selecting the regularization strength from data or for reporting truncation
bias. On the benchmark side, ground-truth trajectory density ratios are almost never available.
analytically tractable diffusions, finite-state Markov chains, and semi-synthetic data can cover
all three levels, but the design must prevent ``the model is correct'' from masquerading as ``the
method works'' \citep{fu2021opebench,voloshin2021empirical,lueckmann2021benchmark,talts2018validating}.
Relative-error guarantees for rare-event estimation, well studied in the Monte Carlo literature
\citep{kahn1953methods,bucklew2004introduction}, have no counterpart for learned trajectory
ratios.

\paragraph{7. From estimates to valid inference.}
The martingale property of equation~\eqref{eq:martingale} is exact for the true prefix ratio but
not inherited by learned estimates. Which constructions, from simulation-based calibration and conservative variance estimates to
betting-based confidence sequences, restore anytime validity, and at what cost
in power, and under which conditional-simulation access, is open
\citep{cortinovis2026slice,forre2026anytime,ramdas2023savi,grunwald2024safe,waudby2024estimating}.
The distinction matters operationally: an estimate whose numerical accuracy has been optimized is
not thereby licensed for sequential decision rules \citep{wald1945sequential,siegmund1985sequential}.

\section{Conclusion}\label{sec:conclusion}

This survey organized density-ratio estimation for process-generated trajectories around a single
claim: the trajectory ratio $w_T$ is the unique estimand that one estimate reuses across every trajectory
functional, weak support makes it nonexistent, uncontrollable, or singular at the three levels
$L_1$--$L_3$, and every method family in the literature is a retreat from this object to a
surrogate estimand with a characteristic profile of what it keeps and what it pays. The two escalation
mechanisms explain why the retreat is forced. Local gaps become trajectory singularity, and thin
overlap amplifies variance exponentially with the horizon. No composite of retreats removes
the pressure, because each retreat pays at a different point: in tails from truncation and
reweighting, in compounding local error from staging and telescoping, in trajectory identity from
stationarization, in point estimates from bounds, in the estimand itself from smoothing,
dequantization, and coupling, and in reusability from bypassing the ratio.

Read column-wise, the map is predictive: a method's known pathologies follow from its retreat, not
from implementation quality, which is why weight clipping biases free energies, why stationarized
estimators miss hitting probabilities, and why transport plans say nothing about ratio values.
Read row-wise, the map is sociological: each application domain standardized the retreat its data
force, and the diagnostics it calls by its own name are the same trigger measured at a different
level. For the newest domain, language-model post-training and multi-turn agents, the map
separates what is known, namely re-leveled sequence ratios with controlled tails and KL anchors
that maintain overlap, from what is open, namely turn-level ratios under unknown user processes and
horizon-aware rates for autoregressive trajectory ratios. The most valuable next step, in our reading, is to
make the retreat itself an object of design: diagnostics that trigger the retreat later, surrogates
that pay less, and rates that say when the surrogate can be trusted as a stand-in for the
trajectory ratio.

%% file: sections/survey_appendix.tex
\section{Technical appendix}\label{app:technical}

This appendix collects the identities used in the main text, with precise conditions and pointers
to the literature. It is a reference tool for the survey's claims, not a source of new results.

\subsection{Variance amplification}\label{app:variance}

\begin{proposition}[Product form of the second moment. Proof of Observation~\ref{obs:amplification}]\label{prop:amplification}
Let $P_T$ and $Q_T$ be the laws of $\rmX_{0:T}$ under the target and data processes, and suppose the
one-step increment ratios $r_t=p_t(\rmX_t\mid \rmX_{t-1})/q_t(\rmX_t\mid \rmX_{t-1})$ are independent and
identically distributed under $Q$. Then
\[
\E_Q\bigl[w_T^2\bigr] \;=\; \prod_{t=1}^{T}\E_Q\bigl[r_t^2\bigr] \;=\; \E_Q\bigl[r_1^2\bigr]^{T},
\]
and $\E_Q[r_1^2]=1+\chi^2(P_1\Vert Q_1)$, with both sides infinite when the $\chi^2$-divergence of
the one-step laws is infinite.
\end{proposition}

\begin{proof}[Proof sketch]
Along the product structure of equation~\eqref{eq:product}, $w_T=\prod_t r_t$. Independence of the
$r_t$ under $Q$ factors the expectation into a product of second moments. Writing
$\E_Q[r^2]=\int (p/q)^2\dd q=\int p^2/q\,\dd q=1+\chi^2(P_1\Vert Q_1)$ gives the identity. Integrability
fails exactly when $\int p^2/q$ diverges \citep{cover2006,tsybakov2009introduction}.
\end{proof}

Under dependence the product form fails, but the exponential rate survives under mixing: for
geometrically ergodic chains the moment generating function of additive functionals is governed by
the spectral radius of a tilted transfer operator \citep{lasota1994chaos,baladi2000positive,lezaud1998chernoff,meyn2009markov},
and for diffusions the Girsanov exponential martingale of equation~\eqref{eq:girsanov} has second
moment $\exp\!\bigl(\int_0^T\|\sigma^{-1}\Delta b\|^2\dd t\bigr)$
\citep{girsanov1960measures,karatzas1991brownian,revuz1999continuous}. The martingale central limit
regime for additive functionals of Markov chains is captured by the Kipnis--Varadhan theory
\citep{kipnis1986convergence,ethier1986markov}. The practical moral is scale-free: truncating the
one-step ratio at level $M$ bounds the trajectory second moment by $M^{2T}$, so per-step tail
control buys exponential amortization \citep{kahn1953methods,owen2000safe,bucklew2004introduction}.

\subsection{The prefix correction identity}\label{app:prefix}

Let $\vh$ denote a history, generated under one of the two mechanisms with probabilities $p(\vh)$ and
$q(\vh)$. By Bayes' rule,
\begin{equation}\label{app:eq:prefix}
\frac{p(\vx\mid \vh)}{q(\vx\mid \vh)}
=\frac{p(\vh,\vx)\,/\,p(\vh)}{q(\vh,\vx)\,/\,q(\vh)}
=\frac{p(\vh,\vx)/q(\vh,\vx)}{p(\vh)/q(\vh)},
\end{equation}
so the conditional one-step ratio is the joint ratio of the extended prefix divided by the joint
ratio of the history. A classifier trained to discriminate $p(\vh,\vx)$ from $q(\vh,\vx)$ therefore estimates
the numerator of equation~\eqref{app:eq:prefix}, and its odds must be normalized by the history
odds before being accumulated into prefix evidence. SPRT-TANDEM's local differencing implements
exactly this normalization \citep{ebihara2021tandem}, and LSEL's consistency analysis requires the
identifiability of the conditional targets \citep{miyagawa2021lsel}. Additive log-scale error
propagation, $\widehat L_T-L_T=\sum_t(\widehat\ell_t-\ell_t)$, follows from
equation~\eqref{eq:chain}. No variance reduction is implied by the logarithm, and the accumulated
error inherits any heavy tails of the local errors \citep{ebihara2023saturation}.

\subsection{Overlap coefficients and divergence relations}\label{app:overlap}

For absolutely continuous $P\ll Q$ with ratio $w=\dd P/\dd Q$,
\[
\varrho(P,Q)=\int\min(w,1)\,\dd Q=1-\TV(P,Q),\qquad
\chi^2(P\Vert Q)=\E_Q[w^2]-1,\qquad
R_2(P\Vert Q)=\log\E_Q[w^2],
\]
so square-integrability of the ratio is exactly finiteness of the R\'enyi-2 divergence, and the
effective sample size of a self-normalized importance sampler is controlled by
$\varrho$ and $\chi^2$ \citep{kahn1953methods,agapiou2017importance,cover2006}. The phase-space
overlap measures of free-energy calculation estimate the same quantities for adjacent windows, with
bias conditions stated in terms of the window ratio's tails \citep{wu2005phase,wu2005relative,pohorille2010}.
Kernel mean embeddings replace these divergences by distances that remain well defined without
absolute continuity, useful precisely when $L_1$ is suspected \citep{gretton2012mmd,sriperumbudur2010hilbert,muandet2017kernel}.

\subsection{Weighted conformal prediction under shift}\label{app:conformal}

Split conformal prediction attains distribution-free coverage under exchangeability
\citep{vovk2005algorithmic,lei2014distribution,angelopoulos2023gentle}. Under covariate shift,
coverage is restored by weighting the conformity scores with the covariate density ratio
\citep{tibshirani2019weightedcp}. For time series, adaptive conformity intervals track drift by
online feedback \citep{gibbs2021aci}, EnbPI constructs approximately marginal bands under strong
mixing of the errors \citep{xu2021enbpi}, and non-exchangeable coverage quantifies the robustness
of conformal validity to distribution drift and asymmetric fitting \citep{barber2023nonexchangeable}.
These are the trajectory-data interfaces of Section~\ref{sec:seq-inference}: the ratio enters as a
weight, and the guarantee is coverage, not point accuracy. If a classifier score whose exponential
is read as a ratio is miscalibrated, it breaks the weighting step silently, which is why calibration
checks precede conformal deployment \citep{guo2017calibration,cranmer2015calibrated}.

\subsection{E-processes and anytime validity}\label{app:anytime}

A non-negative martingale with initial value one is an e-process: by Ville's inequality, the
probability that it ever exceeds $1/\alpha$ is at most $\alpha$, uniformly over stopping times
\citep{ville1939etude,shafer2019game}. The exact prefix ratio of two absolutely continuous
processes is such a martingale by equation~\eqref{eq:martingale}, which is the sequential-analysis
root of early classification \citep{wald1945sequential,wald1948optimal,siegmund1985sequential,lai2001sequential}.
Anytime-valid confidence sequences and betting-based means extend the same discipline to estimation
\citep{howard2021time,waudby2024estimating,shafer2021testing,ramdas2023savi}. For learned scores,
SLICE restores validity by construction: a positive score is selected before the round, calibrated
against fresh simulations under the null, and the normalized e-values multiply into a test
martingale under conditional exchangeability \citep{cortinovis2026slice,forre2026anytime}. The
e-value methodology trades power for this universal stopping-time guarantee
\citep{grunwald2024safe}. The appendix of \citet{cortinovis2026slice} discusses the extension from
i.i.d.\ to conditionally exchangeable, Markov, and state-space simulators, which is the access
condition connecting the construction to Section~\ref{sec:sbi}.

\subsection{Estimation from truncated support}\label{app:truncation}

When observations outside a set are absent, the statistical form of $L_1$, parameter estimation
remains possible with efficient rates when the truncation set is known and geometrically favorable
\citep{daskalakis2018efficient}. Rare-event reliability estimation requires relative rather than
absolute error, and the corresponding sample complexity is governed by the rarity of the event
\citep{karp1983monte,bucklew2004introduction}. Importance-sampling cost is governed by the
$\chi^2$-divergence between target and proposal \citep{agapiou2017importance,chatterjee2018is}.
These results calibrate the expectations of Section~\ref{sec:partial}: inside $L_1$, the achievable
guarantee is a bound with a rate, not a point estimate.

\subsection{Notation}\label{app:notation}

\begin{table}[t]
\centering\small
\caption{Notation used throughout the survey.}
\label{tab:notation}
\begin{tabular}{@{}ll@{}}
\toprule
Symbol & Meaning \\
\midrule
$\gX$, $\Omega_T$, $\gF_T$ & state space; trajectory space $\gX^{T+1}$; trajectory $\sigma$-algebra \\
$\rmX_t$, $\vx_t$ & state at time $t$: random vector and its value \\
$\rmA_t$, $\va_t$ & action at time $t$: random vector and its value \\
$Q$, $P$ & trajectory measures of the data and target processes \\
$q_t$, $p_t$ & one-step transition densities of the data and target processes \\
$\mu$, $\rho$ & initial distributions of the data and target processes \\
$w_T$, $L_T$ & trajectory density ratio $\dd P/\dd Q$ and its logarithm \\
$r_t$ & one-step increment ratio $p_t/q_t$ \\
$\zeta$ & stationary (state--action) density ratio $\dd\pi/\dd D$ \\
$\TV$, $\chi^2$, $R_\alpha$ & total variation, $\chi^2$-divergence, R\'enyi-$\alpha$ divergence \\
$\varrho(P,Q)$ & overlap coefficient $1-\TV(P,Q)$ \\
$\esssup$ & essential supremum with respect to the ambient measure \\
$d$, $D$ & occupancy measure of the target policy; stationary distribution of the data policy \\
$e_t$ & e-value at round $t$ (non-negative, unit mean under the null) \\
\bottomrule
\end{tabular}
\end{table}

%% file: sections/survey_retreatmap.tex
\section{The map of retreats on one page}\label{app:retreats}

Table~\ref{tab:retreats} states the organizing claim of this survey on one page: every method in
the corpus retreats from the trajectory ratio $w_T$ to one of nine surrogate estimands. The four
tick columns record what each surrogate keeps, and an unchecked column is exactly what it pays.
The level column names the weak-support level that forces the retreat, and the last columns name
the standardizing domains and a key work. The classification is by \emph{primary} retreat. Deployed
systems compose several.

\begingroup
\footnotesize
\setlength{\tabcolsep}{3.5pt}
\renewcommand{\arraystretch}{1.0}
\begin{table}[ht]
\centering\footnotesize
\caption{The nine retreats and what each preserves
(\checkmark\ = kept, $\circ$ = partly, --- = given up).}\label{tab:retreats}
\begin{tabular}{@{}>{\raggedright\arraybackslash}p{26mm} >{\centering\arraybackslash}p{11mm} >{\centering\arraybackslash}p{9mm} >{\centering\arraybackslash}p{11mm} >{\centering\arraybackslash}p{13mm} >{\centering\arraybackslash}p{8mm} >{\raggedright\arraybackslash}p{37mm} >{\raggedright\arraybackslash}p{26mm}@{}}
\toprule
Retreat & Traj.\ detail & Reuse & Point est. & Horizon-free & Level & Standardized by & Key work \\
\midrule
Direct estimation (\S\ref{sec:kernels}) &
\checkmark & \checkmark & \checkmark & $\times$ & --- &
known-probability reweighting, classification &
\citet{kanamori2009ulsif} \\
\addlinespace[1pt]
Truncation and reweighting (\S\ref{sec:reweighting}) &
\checkmark & \checkmark & \checkmark & $\times$ & $L_2$ &
particle filters, alchemical windows, policy lag &
\citet{kahn1953methods} \\
\addlinespace[1pt]
Staging and telescoping (\S\ref{sec:sequence}) &
$\circ$ & $\circ$ & \checkmark & $\times$ & $L_2$ &
WHAM staging, prefixes, compositional SBI &
\citet{kumar1992weighted} \\
\addlinespace[1pt]
Stationarization (\S\ref{sec:dice}) &
-- & $\times$ & \checkmark & \checkmark & $L_2$ &
DICE family, ecology RSFs &
\citet{liu2018horizon} \\
\addlinespace[1pt]
Bounds (partial identification) (\S\ref{sec:partial}) &
-- & $\times$ & $\times$ & \checkmark & $L_1$ &
pessimism, attribution, selection &
\citet{khan2023off} \\
\addlinespace[1pt]
Smoothing and dequantization (\S\ref{sec:bridges}) &
\checkmark & \checkmark & \checkmark & \checkmark & $L_3$ &
dequantized diffusion--Schr\"odinger bridges &
\citet{chen2025d3re} \\
\addlinespace[1pt]
Coupling by transport (\S\ref{sec:bridges}) &
-- & $\times$ & $\times$ & \checkmark & $L_3$ &
entropic OT, single-cell responses &
\citet{cuturi2013sinkhorn} \\
\addlinespace[1pt]
Operator spectra (\S\ref{sec:operators}) &
-- & $\times$ & $\times$ & \checkmark & $L_3$ &
MSMs, coherent sets &
\citet{koopman1931hamiltonian} \\
\addlinespace[1pt]
Bypassing the ratio (\S\ref{sec:bypass}) &
-- & $\times$ & \checkmark & \checkmark & any &
DR-OPE, model substitution &
\citet{jiang2016dr} \\
\bottomrule
\end{tabular}
\end{table}
\endgroup

Two readings of the table are worth making explicit. Read column-wise, the unchecked boxes are
the survey's predictive claim: a method's known failure modes are exactly the properties its
retreat gives up, which is why weight clipping biases free energies, why stationarized
estimators miss hitting probabilities, and why transport plans say nothing about ratio values. Read
row-wise, the ``standardized by'' column is the survey's sociological claim: each discipline
converged on the retreat its data force, and the diagnostics it calls by its own name
(phase-space overlap, effective sample size, positivity, concentrability) are all the same
trigger measured at a different level.